\documentclass{article}

\usepackage[preprint]{neurips_2026}

\usepackage[utf8]{inputenc} 
\usepackage[T1]{fontenc}    
\usepackage{hyperref}       
\usepackage{url}            
\usepackage{booktabs}       
\usepackage{amsfonts}       
\usepackage{nicefrac}       
\usepackage{microtype}      
\usepackage{xcolor}         

\usepackage{algorithm}
\usepackage{algpseudocode}

\usepackage{arydshln}   
\usepackage{algpseudocode}
\usepackage{times}
\usepackage{latexsym}
\usepackage{amsmath}
\usepackage{amssymb}
\usepackage{booktabs} 
\usepackage{enumitem}
\usepackage{graphicx}
\usepackage{pifont}
\usepackage{color}
\usepackage{titletoc}
\usepackage{colortbl}
\usepackage{wrapfig}  
\usepackage{lipsum} 
\usepackage{listings}
\usepackage{array}
\usepackage{siunitx}
\usepackage{tcolorbox}
\usepackage{booktabs}
\usepackage{tabularx}
\usepackage{caption}
\usepackage{pifont}
\usepackage{diagbox}
\usepackage{multirow}
\usepackage{soul}
\usepackage{natbib}

\newcommand{\circled}[1]{\ding{\numexpr171+#1\relax}} 
\definecolor{gptgray}{RGB}{245,246,248}
\definecolor{gptbar}{RGB}{230,232,236}
\definecolor{qwenlav}{RGB}{244,240,252}
\definecolor{qwenbar}{RGB}{228,220,246}
\definecolor{glmblue}{RGB}{235,242,250}
\newcommand{\gain}[1]{\textcolor{blue}{\makebox[1.8em][r]{\tiny(#1)}}}
\newcommand{\drop}[1]{\textcolor{red}{\makebox[1.8em][r]{\tiny(#1)}}}

\usepackage{graphicx}

\definecolor{red}{RGB}{255,0,0}
\definecolor{blue}{RGB}{0,0,255}
\definecolor{light-green}{RGB}{220,255,220}
\definecolor{light-blue}{cmyk}{0.1,0,0,0}
\definecolor{officialbg}{HTML}{FFF2CA}
\definecolor{opensourcebg}{HTML}{E9DCEE}
\definecolor{officialbgsub}{HTML}{FFF8E3}
\definecolor{opensourcebgsub}{HTML}{F4ECF7}

\title{Connecting the Dots: Benchmarking Reflective Memory in Long-Horizon Dialogue}

\author{Jingjie Lin$^{1*}$, Bingbing Wang$^{1,2*}$,Zihan Wang$^{1}$, Zhengda Jin$^{1}$, Weiming Qiao$^{3}$ \\ \bf Jing Li$^{2}$, Ruifeng Xu$^{1,4\dag}$ \\
$^{1}$ Harbin Institute of Technology, Shenzhen, $^{2}$ The Hong Kong Polytechnic University, \\  $^{3}$ Fudan University
, $^{4}$ Peng Cheng Laboratory, bingbing.wang@stu.hit.edu.cn}

\begin{document}

\maketitle

\begin{abstract}
Despite substantial progress in long-context modeling, existing benchmarks remain confined to factual memory for explicit recall, failing to measure the reflective memory required to synthesize fragmented, multimodal cues into high-level interpretations. 
To address this gap, we introduce \textbf{RefMem-Bench}, a benchmark for reflective memory in long-horizon dialogue. 
RefMem-Bench contains 26K annotated QA instances with eight reflective-memory dimensions and three task formats, requiring
models to move beyond surface-level retrieval and infer latent meanings from evidence
distributed across interaction histories. 
To enhance reflective memory capability, we propose \textbf{RE}flective \textbf{M}emory \textbf{IND}uction (\textbf{REMIND}),
a hierarchical framework that treats reflective memory as progressive meaning construction. 
REMIND couples question-conditioned evidence retrieval, salience-aware grounding, and abstraction-level supervision, and uses Progressive Reflective
Alignment to distill high-level reflective reasoning into the factual inference pathway. 
Experiments show RefMem-Bench poses a substantial challenge to current models, while REMIND consistently improves both answer accuracy and memory recall through progressive evidence perception, grounding, and abstraction.
\end{abstract}

\section{Introduction}

As Large Language Models (LLMs) are applied to increasingly long and multimodal contexts, memory has emerged as a key factor in their practical performance \citep{lewis2020retrieval, zhong2024memorybank}. A strong model must do more than process the immediate input by instead incrementally accumulating user experiences, tracking evolving states, and capturing recurring behavioral patterns across extended turns and sessions \citep{wang2023augmenting}. Recent advances in retrieval-augmented generation, memory-augmented modeling, and long-context training have significantly improved the ability of LLMs to access vast interaction histories \citep{team2024gemini, bulatov2023scaling}. These developments move LLMs closer to becoming persistent companions that can leverage a wealth of prior experiences to produce responses that are both personalized and contextually grounded.

\begin{figure}[!t]
  \centering
  \includegraphics[width=0.95\linewidth]{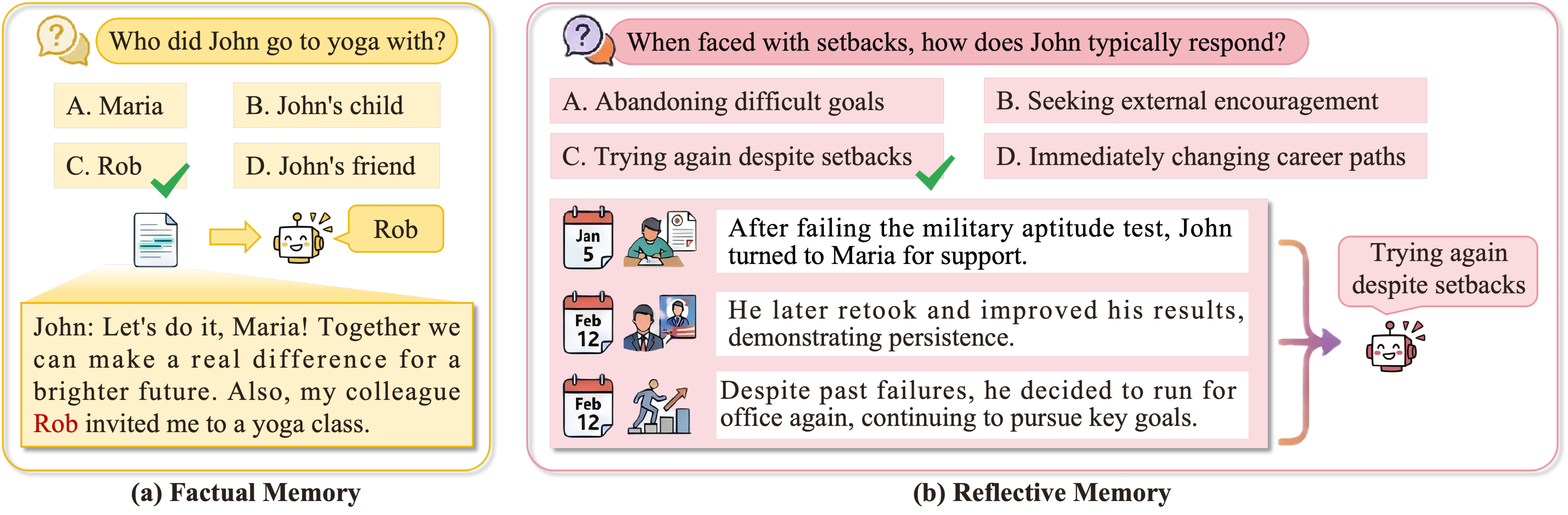}
  \caption{Example of factual memory and reflective memory.}
  \label{fig:intro}
  \vspace{-8pt}
\end{figure}

However, existing evaluation paradigms are still largely centered on explicit memory recall, measuring whether a model can retrieve information that has been directly stated in the interaction history. Benchmarks such as LongBench \citep{bai2025longbench}, LoCoMo \citep{maharana2024evaluating}, and LongMemEval \citep{wu2024longmemeval} are representative examples of this setting. As illustrated in Figure~\ref{fig:intro} (a), these tasks typically require models to identify surface-level details from prior context, such as a specific name or a visually observable object. While such abilities are important, they capture only a limited notion of memory. From a schema-based view of memory \citep{alba1983memory}, effective remembering involves not only retaining details, but also synthesizing distributed signals into structured, higher-level representations. In long-horizon dialogue, this means that \textit{models must go beyond explicit retrieval and integrate scattered historical cues to infer latent states, behavioral patterns, or personal traits}, as illustrated in Figure~\ref{fig:intro} (b). Yet current benchmarks provide little systematic evaluation of this higher-level synthesis, creating the need for a dedicated benchmark for reflective memory that is essential to building truly intelligent long-term agents.

To bridge this gap, we introduce \textbf{RefMem-Bench}, a benchmark for reflective memory in long-horizon dialogue. 
RefMem-Bench contains 26K annotated question-answer instances covering 1K sessions and 71K dialogue turns. Unlike benchmarks centered on explicit recall, it spans eight reflective-memory dimensions and three task formats, requiring models to integrate distributed evidence across extended interaction histories to infer latent user states, behavioral regularities, temporal evolution, and implicit constraints. All instances are evidence-anchored and human-verified to support reliable evaluation of memory reasoning.
To improve performance on this capability, we further propose \textbf{REMIND} (\textbf{RE}flective \textbf{M}emory \textbf{IND}uction), 
a hierarchical framework that treats reflective memory as meaning construction rather than evidence retrieval alone. REMIND organizes reasoning through a three-level \textbf{Cognitive Pyramid}, consisting of factual, attentional, and reflective states, and is trained with \textbf{Progressive Reflective Alignment} to distill high-level reflective reasoning into the factual inference pathway, enabling efficient test-time inference without explicit multi-stage reasoning.

To summarize, our work provides three key contributions: 1) We formalize {reflective memory} as a distinct target for long-term LLM evaluation, introduce {RefMem-Bench} as a large-scale multimodal benchmark with evidence-anchored and human-verified annotation; 2) we propose {REMIND} as a unified framework that integrates factual retrieval, salience grounding, and abstraction induction in a hierarchical training objective; and 3) extensive experiments show that RefMem-Bench is highly challenging for current models, while REMIND consistently improves both answer accuracy and memory recall across task formats and backbone settings.

\section{Related Work}
\textbf{Long-Term Dialogue Benchmarks.}
Recent long-term dialogue benchmarks have substantially advanced the evaluation of memory in sustained interactions, yet most remain centered on explicit recall. Representative examples include LoCoMo~\citep{maharana2024evaluating} and LongMemEval~\citep{wu2024longmemeval}, which evaluate memory through long-horizon question answering, summarization, temporal reasoning, and knowledge updates. REALTALK~\citep{lee2025realtalk}, PersonaMem~\citep{jiang2025know}, and LifeBench~\citep{cheng2026lifebench} move toward more realistic long-span interactions and evolving user profiles, while MADial~\citep{he2025madial}, MMRC~\citep{xue2025mmrc}, and MemGallery~\citep{bei2026mem} extend evaluation to memory-aware generation and multimodal conversation. In contrast, our work focuses on reflective memory: whether models can integrate temporally scattered and cross-modal cues into coherent latent interpretations.


\textbf{Long-Term Memory Methods.}
Existing long-term memory methods for LLMs mainly improve interaction over long horizons by introducing external memory structures, latent memory mechanisms, or adaptive retrieval pipelines. Early systems such as MemoryBank~\citep{zhong2024memorybank} and MemGPT~\citep{packer2023memgpt} show that explicit memory storage can help preserve salient user information beyond the immediate context window, while model-level approaches such as LongMem~\citep{wang2023augmenting} and MemoryLLM~\citep{wang2024memoryllm} explore cached context and latent memory pools for scalable long-range retention. More recent systems, including Mem0~\citep{chhikara2025mem0}, MemoryOS~\citep{kang2025memory}, A-Mem~\citep{xu2025mem}, and M+~\citep{wang2025m+}, further improve scalability, structured storage, retrieval quality, and adaptive memory management. In contrast to the above memory systems, we treat long-term memory as hierarchical meaning construction, emphasizing how models transform retrieved traces into abstraction-level reasoning.

\begin{figure}[!t]
  \centering
  \includegraphics[width=\linewidth]{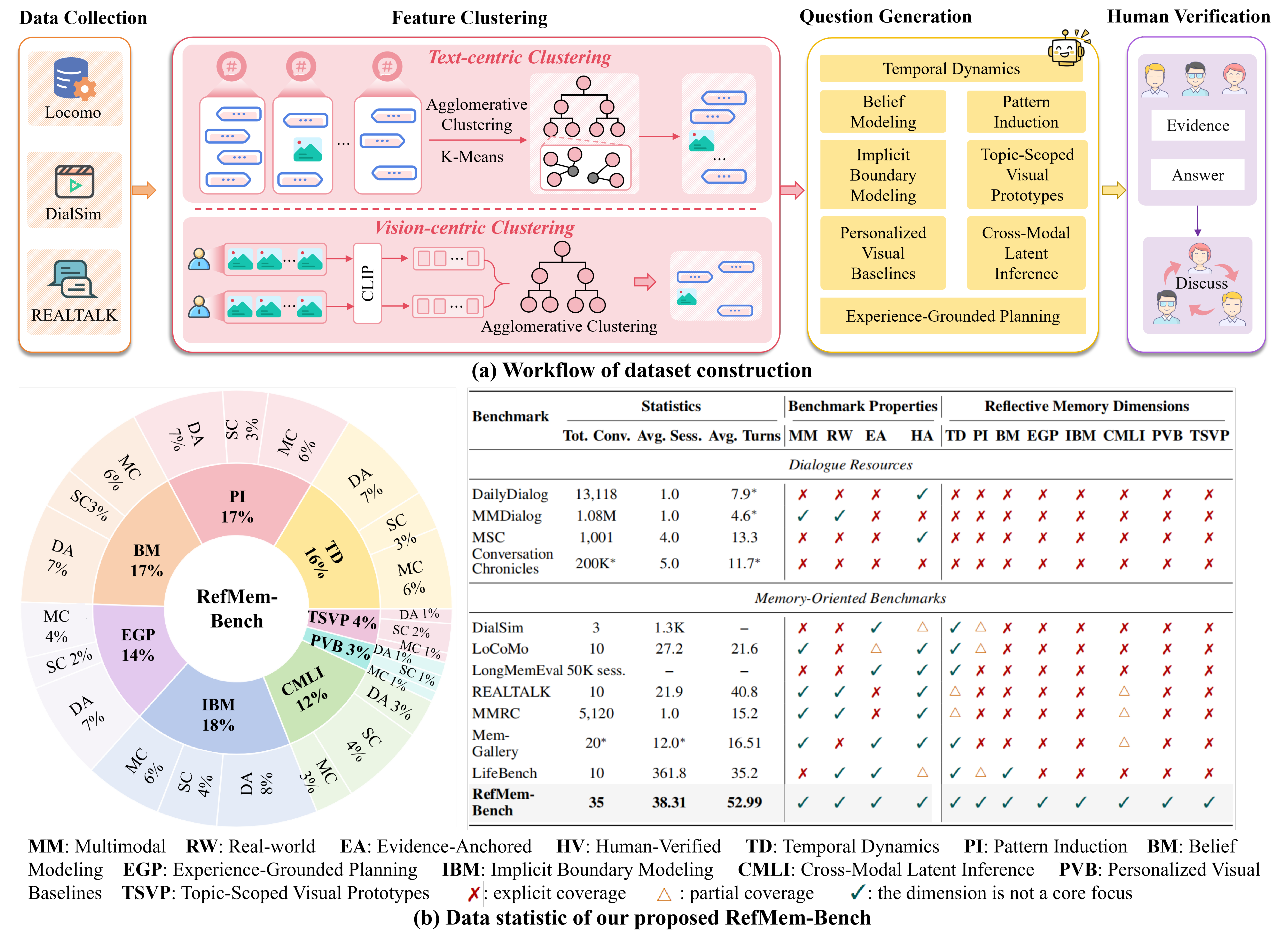}
  \caption{Overview of RefMem-Bench. (a) Workflow of dataset construction. (b) 
  Comparison between RefMem-Bench and representative dialogue datasets and memory benchmarks. Statistics marked with $^{*}$ are derived from values reported in the original papers.}
  \label{fig:dataset_statistic}
  \vspace{-5pt}
\end{figure}

\section{Benchmark}
\label{sec:benchmark}
\textbf{Dataset Curation.}
RefMem-Bench is built through a semi-automatic pipeline over long-horizon conversations collected from REALTALK~\citep{lee2025realtalk}, DialSim~\citep{kim2024dialsim}, and LoCoMo~\citep{maharana2024evaluating}, with complementary text-centric and image-centric branches (Figure~\ref{fig:dataset_statistic}). In the text branch, we segment multi-turn dialogues into fine-grained topical units, cluster them with Agglomerative Clustering \citep{johnson1967hierarchical}, and refine large clusters via K-Means \citep{macqueen1967some}, producing questions for \textbf{Temporal Dynamics}, \textbf{Pattern Induction}, \textbf{Belief Modeling}, \textbf{Experience-Grounded Planning}, and \textbf{Implicit Boundary Modeling}. In the image branch, we collect speaker-level image histories from REALTALK, encode them with CLIP \citep{radford2021learning}, and cluster them to construct questions for \textbf{Cross-Modal Latent Inference}, \textbf{Personalized Visual Baselines}, and \textbf{Topic-Scoped Visual Prototypes}. 
All generated instances are reviewed by three expert annotators to verify answers and evidence, correct LLM-induced errors, and filter out ambiguous or low-quality questions, with inconsistent cases resolved by an adjudicator following prior annotation protocols~\citep{wu2024longmemeval}. More details are provided in Appendix~\ref{app:dataset_curation}.

\textbf{Task Formulation.}
We instantiate this formulation with three complementary task formats: \textbf{single-choice} (\textbf{SC}), where the model selects one answer from four candidates; \textbf{multiple-choice} (\textbf{MC}), where it identifies all valid options among four candidates; and \textbf{direct-answer} (\textbf{DA}), where it produces a short free-form response that is matched against a small set of acceptable references. 

\noindent\textbf{Data Statistics.}
RefMem-Bench contains 26,449 annotated QA instances constructed from 35 curated long-horizon conversational contexts. Each instance includes a complete conversational thread and, when available, aligned visual evidence. Figure~\ref{fig:dataset_statistic} shows that RefMem-Bench is substantially larger and more comprehensive than prior dialogue datasets and memory benchmarks in reflective-memory coverage. To the best of our knowledge, it is the only benchmark in our comparison that simultaneously supports multimodal, real-world, evidence-anchored, and human-verified evaluation across all eight reflective-memory dimensions. Following recent long-horizon memory-agent benchmarks, including Mem0~\citep{chhikara2025mem0} and Memory-R1~\citep{yan2025memory}, we use a 2:8 train/test split. Detailed statistics by task format and reflective-memory dimension are provided in Appendix~\ref{app:data_statistic}.

\textbf{Evaluation Protocols.} Following \citet{maharana2024evaluating} and \citet{wu2024longmemeval}, we evaluate all task formats using two primary metrics, answer accuracy (\textbf{Acc}) and memory recall (\textbf{MemR}), and additionally report BLEU-1 (\textbf{B1}) and token-level F1 (\textbf{F1}) for direct-answer (DA) questions. For SC and MC, Acc is computed by exact match between predicted and gold option sets; for DA, correctness is determined by an LLM-as-a-judge~\citep{yang2023gpteval} using \texttt{gpt-4o-2024-08-06}. 
MemR is unified across the benchmark with modality-specific scoring: semantic evidence matching for text-dominant dimensions and evidence index overlap for visually grounded dimensions, both averaged over valid samples. For DA, B1 and F1 are computed from unigram overlap between predictions and references. Further details are provided in Appendices~\ref{app:dataset_evaluation} and~\ref{app:meta_experiments}.

\section{REMIND Method}
Given a long-horizon multimodal dialogue history $\mathcal{H}$ and a reflective-memory question $q$, our goal is to generate an answer $y$. Unlike standard memory QA, reflective-memory questions require both evidence retrieval and higher-level inference over clues distributed across turns, sessions, and modalities. To address this challenge, we propose \textbf{RE}flective \textbf{M}emory \textbf{IND}uction (\textbf{REMIND}), a hierarchical framework with a three-level Cognitive Pyramid, trained via Progressive Reflective Alignment to distill higher-level reasoning into the factual inference pathway used at test time.

\begin{figure}[!t]
  \centering
\includegraphics[width=0.9\linewidth]{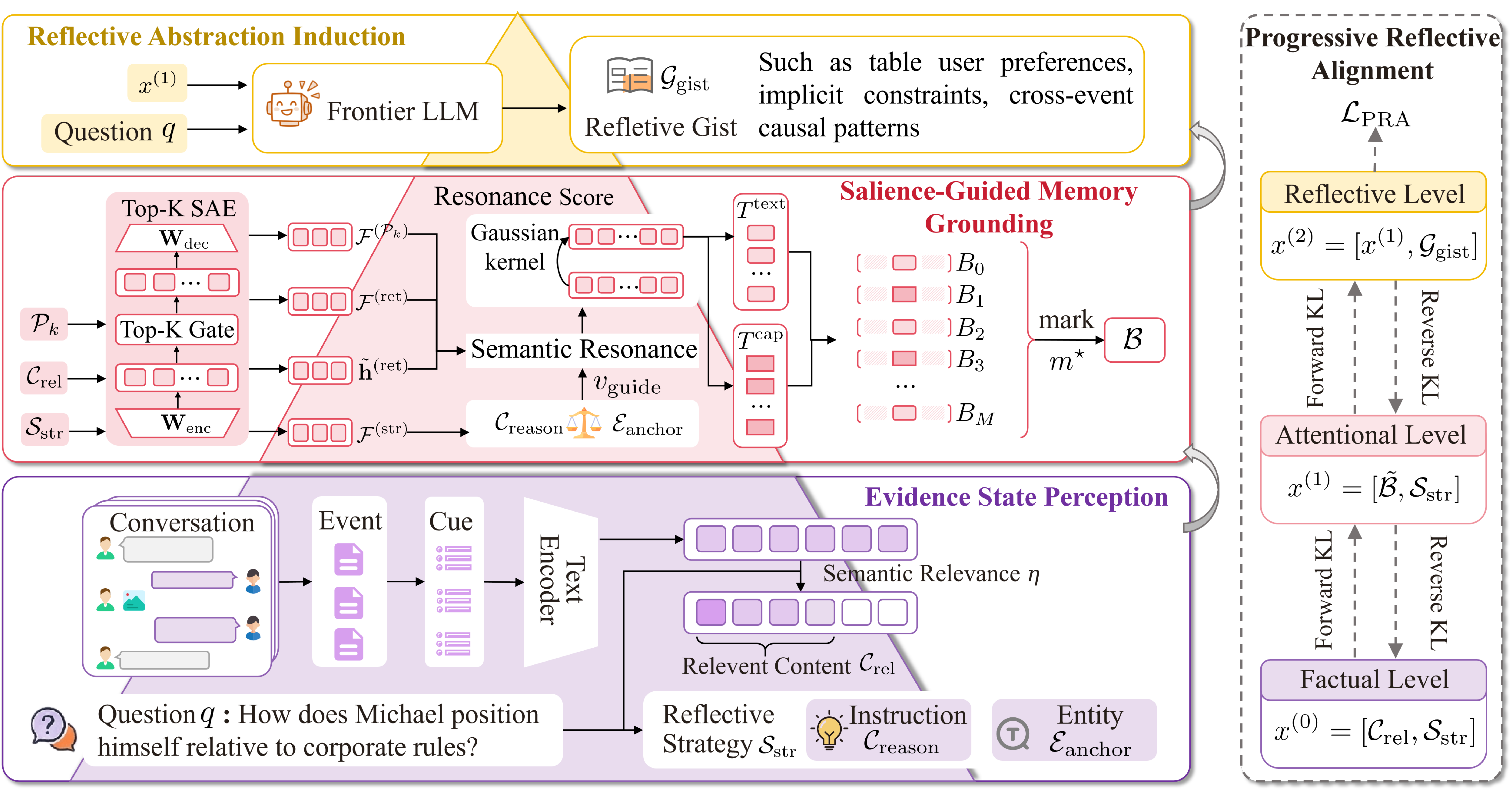}
  \caption{Overview of our REMIND method.
}
  \label{fig:method}
  \vspace{-5pt}
\end{figure}

\subsection{Cognitive Pyramid Modeling}
Reflective reasoning in long-horizon dialogue is inherently hierarchical: answering a reflective-memory question requires retrieving relevant evidence, identifying salient clues, and synthesizing them into higher-level interpretations. Consistent with a schema-based view of memory \citep{alba1983memory} and prior work on hierarchical cognitive control \citep{badre2008cognitive}, we formulate this process as a three-level \textbf{Cognitive Pyramid} with factual, attentional, and reflective levels. As shown in Figure~\ref{fig:method}, these levels correspond to \emph{Evidence State Perception}, \emph{Salience-Guided Memory Grounding}, and \emph{Reflective Abstraction Induction}. Finally, \textbf{Progressive Reflective Alignment} distills higher-level reasoning into the factual inference pathway.

Let $x^{(i)}$ denote the state at level $i$, where $x^{(0)}$, $x^{(1)}$, and $x^{(2)}$ correspond to the factual, attentional, and reflective states, respectively. Given $x^{(i)}$, the model defines the autoregressive next-token distribution
$p_i(y_t \mid y_{<t})
\equiv
p_{\theta}(y_t \mid x^{(i)}, y_{<t})$,
where $y_{<t}=(y_1,\ldots,y_{t-1})$ and $\theta$ denotes the model parameters.

The central idea of REMIND is to model reflective memory as progressive meaning construction rather than evidence retrieval alone. During training, the distribution induced by the factual state is encouraged to approximate those induced by higher-level states, so that attentional grounding and reflective abstraction are distilled into the base inference pathway. This yields efficient test-time reasoning without explicit multi-stage execution.

\subsection{Evidence State Perception}

We first instantiate the \emph{factual level} by constructing a question-conditioned factual state $x^{(0)}$. This stage couples coarse-grained evidence retrieval with a structured reflective strategy, yielding broad but high-signal memory access for downstream grounding and abstraction. Specifically, we decompose the question into 
$\mathcal{S}_{\mathrm{str}}
=
\bigl(\mathcal{C}_{\mathrm{reason}}, \mathcal{E}_{\mathrm{anchor}}\bigr)$,
where $\mathcal{C}_{\mathrm{reason}}$ is a set of generic reasoning instructions and $\mathcal{E}_{\mathrm{anchor}}$ is a set of explicit answer-critical anchors, such as entities, attributes, or concepts.

Because the full dialogue history $\mathcal{H}$ contains substantial redundancy and multimodal noise, we first partition it into $N$ temporally ordered segments:
$\mathcal{U}=\{u_n\}_{n=1}^{N}$.
For each segment $u_n$, we derive a segment-level event cue $\kappa_n$ tailored to the question type and its cognitive demand from LLM. We then encode both the question and the cue with a text encoder $\phi(\cdot)$ and compute a relevance score $\eta_n$ that measures the semantic relevance between question $q$ and segment cue $\kappa_n$:
\begin{equation}
\eta_n
=
\operatorname{sim}(q,\kappa_n)
=
\tfrac{\phi(q)^{\top}\phi(\kappa_n)}
{\|\phi(q)\|_2\,\|\phi(\kappa_n)\|_2}.
\end{equation}
We rank candidate segments by $\eta_n$ and retain the top-$k_1$ as $\mathcal{C}_{\mathrm{ret}}$, yielding the factual state
$x^{(0)}=[\mathcal{C}_{\mathrm{ret}}, \mathcal{S}_{\mathrm{str}}]$.
This question-conditioned retrieval filters irrelevant events while preserving broad evidence coverage for subsequent grounding and abstraction.

\subsection{Salience-Guided Memory Grounding}

Starting from the coarse factual state $x^{(0)}$, this section highlights strategy-salient evidence and preserves its local interaction structure, yielding the attentional state $x^{(1)}$ for downstream abstraction.

\textbf{Sparse Encoding.}
To enable strategy-guided grounding, we map both the reflective strategy and the retrieved context into a shared sparse latent space using a Top-K SAE~\citep{gao2024scaling}. This sparse representation highlights the most salient semantic features of each token, allowing strategy and context tokens to be compared at the feature level rather than only through surface lexical overlap.

Let $\mathbf{h}_t^{(x)} \in \mathbb{R}^{d_{\mathrm{model}}}$ denote the hidden state of token $t$, where $x \in \{\mathrm{str}, \mathrm{ret}\}$ indicates whether the token belongs to the reflective strategy or the retrieved context. 
The Top-K SAE maps $\mathbf{h}_t^{(x)}$ to a sparse latent representation $\tilde{\mathbf{h}}_t^{(x)} \in \mathbb{R}^{d_{\mathrm{latent}}}$ and active feature set $\mathcal{F}_t^{(x)}$:
\begin{equation}
\tilde{\mathbf{h}}_t^{(x)}
=
\operatorname{TopK}\big(
\operatorname{ReLU}(
\mathbf{W}_{\mathrm{enc}}(\mathbf{h}_t^{(x)}-\mathbf{b}_{\mathrm{dec}})
+\mathbf{b}_{\mathrm{enc}}
),\, k
\big),
\qquad
\mathcal{F}_t^{(x)}
=
\{j \mid [\tilde{\mathbf{h}}_t^{(x)}]_j > 0\}.
\end{equation}
Here, $\mathbf{W}_{\mathrm{enc}} \in \mathbb{R}^{d_{\mathrm{latent}} \times d_{\mathrm{model}}}$ and $\mathbf{b}_{\mathrm{enc}} \in \mathbb{R}^{d_{\mathrm{latent}}}$ are the encoder parameters, $\mathbf{b}_{\mathrm{dec}} \in \mathbb{R}^{d_{\mathrm{model}}}$ is the decoder bias, $\operatorname{ReLU}(\cdot)$ is the rectified linear unit, and $\operatorname{TopK}(\cdot,k)$ keeps the top-$k_1$ activated dimensions and zeros out the rest.

\textbf{Guidance Projection.}
We next project the reflective strategy $\mathcal{S}_{\mathrm{str}}$ into the same sparse space to obtain a global guidance vector. Specifically, we aggregate feature activations from reasoning spans and anchor spans, while downweighting generic features and emphasizing answer-critical ones:
\begin{equation}\small
[\mathbf{v}_{\mathrm{guide}}]_j
=
\max\Big(
w_{\mathrm{idf}}(j)
\sum_{t \in \mathcal{T}_{\mathrm{reason}}}
\mathbb{I}[j \in \mathcal{F}_t^{(\mathrm{str})}],
\;
\alpha
\sum_{t \in \mathcal{T}_{\mathrm{anchor}}}
\mathbb{I}[j \in \mathcal{F}_t^{(\mathrm{str})}]
\Big).
\end{equation}
Here, $\mathcal{T}_{\mathrm{reason}}$ and $\mathcal{T}_{\mathrm{anchor}}$ denote reasoning-span and anchor-span token sets from $\mathcal{C}_{\mathrm{reason}}, \mathcal{E}_{\mathrm{anchor}}$, $w_{\mathrm{idf}}(j)$ is the inverse-document-frequency weight of latent feature $j$, and $\alpha$ controls anchor boosting. The resulting vector $\mathbf{v}_{\mathrm{guide}}$ serves as a compact descriptor of the reflective strategy.

\textbf{Semantic Resonance.}
Coarse retrieval provides broad coverage but remains noisy for reflective grounding, since salient clues are often sparse and embedded in local interaction structure. We therefore refine the retrieved dialogue by measuring token-level alignment with the strategy guidance vector. For each turn $T_k \in \mathcal{C}_{\mathrm{ret}}$, let $\mathcal{P}_k$ and $\mathcal{F}^{(\mathcal{P}_k)}$ denote its speaker and the feature set of its speaker marker, respectively. The token-level resonance score is defined as
\begin{equation}\small
r_t
=
\underbrace{
\sum_{j \in \mathcal{F}_t^{(\mathrm{ret})}}
[\mathbf{v}_{\mathrm{guide}}]_j
[\tilde{\mathbf{h}}_t^{(\mathrm{ret})}]_j
}_{\text{local alignment}}
+
\underbrace{
\beta
\sum_{k=1}^{M}
\mathbb{I}[t \in T_k \cup T_{k-1}]
\sum_{j \in \mathcal{F}_t^{(\mathcal{P}_k)}}
[\mathbf{v}_{\mathrm{guide}}]_j
}_{\text{speaker-conditioned broadcast}},
\end{equation}
where $\beta$ controls turn-level broadcasting and $T_0=\varnothing$. The first term captures direct alignment between a token and the reflective strategy in sparse feature space, while the second propagates speaker-conditioned salience to the current turn and its immediate predecessor, helping recover triggering context for replies whose relevance is implicit.

\textbf{Turn Selection.} To obtain a more stable estimate of local semantic density, we smooth token-level resonance scores with a one-dimensional Gaussian kernel $\mathcal{K}$:
$
s_t = r_t \ast \mathcal{K}.
$
For multimodal dialogue, we partition each turn into a text subset $T_k^{\mathrm{text}}$ and a visual-description subset $T_k^{\mathrm{cap}}$, and compute turn salience via weighted fusion:
\begin{equation}\small
s_{\mathrm{turn}}(T_k)
=
(1-\lambda)\,
\operatorname{Agg}\!\bigl(\{s_t \mid t \in T_k^{\mathrm{text}}\}\bigr)
+
\lambda\,
\operatorname{Agg}\!\bigl(\{s_t \mid t \in T_k^{\mathrm{cap}}\}\bigr),
\end{equation}
where $\lambda \in [0,1]$ controls modality fusion and $\operatorname{Agg}(\cdot)$ maps token-level salience to subset-level scores.

We retain salient turns
$
\mathcal{T}_{\mathrm{core}}
=
\{T_k \in \mathcal{C}_{\mathrm{ret}} \mid s_{\mathrm{turn}}(T_k) > \delta\},
$
where $\delta$ is a salience threshold, and expand each selected turn with a neighborhood window of size $w$ to obtain
$
\mathcal{T}_{\mathrm{final}}
=
\bigcup_{T_k \in \mathcal{T}_{\mathrm{core}}}
\{T_{k-w}, \ldots, T_{k+w}\}.
$
Adjacent turns in $\mathcal{T}_{\mathrm{final}}$ are then merged into contiguous semantic blocks
$
\mathcal{B}=\{B_i\}_{i=1}^{M},
\quad
B_i=[T_{s_i}, \ldots, T_{e_i}],
$
where $s_i$ and $e_i$ denote the start and end turn indices of block $B_i$, respectively. This preserves salient evidence together with the local conversational context required for coherent interpretation.

Rather than discarding non-selected context, we preserve the full sequence and inject soft boundary markers to guide attention. For each block $B_i \in \mathcal{B}$, we insert a special marker $m^\star$ at its boundaries:
$
\mathcal{C}_{\mathrm{anchor}}
=
\dots \oplus m^\star \oplus B_i \oplus m^\star \oplus \dots.
$
The final attentional state is then
$
x^{(1)}=[\mathcal{C}_{\mathrm{anchor}}, \mathcal{S}_{\mathrm{str}}].
$
These markers softly bias attention toward strategy-aligned evidence while preserving surrounding context, allowing downstream reasoning to access both salient spans and their triggering structure.

\subsection{Reflective Abstraction Induction}
While reflective grounding specifies \emph{where} the model should attend, the reflective state must also capture \emph{what} the selected evidence jointly implies. To provide high-level supervision during training, we generate an auxiliary reflective summary using a proprietary frontier LLM:
$\mathcal{G}_{\mathrm{ref}} = \mathrm{LLM}(q, x^{(1)})$.
Appending this summary to the grounded input yields the third-level state
$x^{(2)} = [x^{(1)}, \mathcal{G}_{\mathrm{ref}}]$.
$\mathcal{G}_{\mathrm{ref}}$ serves as a compact semantic target that captures the latent structure of the grounded evidence, such as stable user preferences, implicit constraints, or cross-event causal patterns. It is used only during training and is omitted at inference time.

\subsection{Progressive Reflective Alignment}
Given the factual, attentional, and reflective states $x^{(0)}$, $x^{(1)}$, and $x^{(2)}$, we align adjacent levels of the Cognitive Pyramid via progressive distillation. The goal is to transfer attentional grounding and reflective reasoning from higher-level states into the factual pathway rooted in $x^{(0)}$, so that inference requires only the factual state.
Let $p_i^\ell$ denote the temperature-scaled output distribution at output position $\ell$ under state $x^{(i)}$, where $L_y$ is the target sequence length. For each adjacent transition $i \rightarrow i+1$, the bidirectional distillation loss is defined as
{\small
\begin{equation}
\mathcal{L}_{\mathrm{dist}}^{(i \rightarrow i+1)}
=
\frac{1}{L_y}\sum_{\ell=1}^{L_y}
\left[
\alpha_i \, \mathrm{KL}(p_{i+1}^\ell \,\|\, p_i^\ell)
+
(1-\alpha_i)\, \mathrm{KL}(p_i^\ell \,\|\, p_{i+1}^\ell)
\right]
\end{equation}
}
where $\alpha_i \in [0,1]$ weights the forward and reverse KL terms.
To preserve factual accuracy, we further impose hard supervision on the factual state. The overall PRA objective is
{\small
\begin{equation}
\begin{aligned}
\mathcal{L}_{\mathrm{hard}}
=
-\frac{1}{L_y}
\sum_{\ell=1}^{L_y}
\log p_0(y_\ell^{*} \mid x^{(0)}, y_{<\ell}^{*}), ~~
\mathcal{L}_{\mathrm{PRA}}
=
\lambda_{\mathrm{hard}} \mathcal{L}_{\mathrm{hard}}
+
(1-\lambda_{\mathrm{hard}})\tau^2
\sum_{i=0}^{1}
\gamma_i \mathcal{L}_{\mathrm{dist}}^{(i \rightarrow i+1)}
\end{aligned}
\end{equation}
}
where $y_\ell^{*}$ denotes the ground-truth token at position $\ell$, $\lambda_{\mathrm{hard}}$ balances hard supervision and soft distillation, and $\gamma_i$ controls the contribution of each pyramid transition.
By minimizing $\mathcal{L}_{\mathrm{PRA}}$, REMIND progressively distills attentional grounding from $x^{(1)}$ and reflective abstraction from $x^{(2)}$ into the factual pathway $x^{(0)}$, enabling efficient inference using only $x^{(0)}$ at test time.

\section{Experiment}

\subsection{Experimental Setup} 
\label{sec:experiment_setup}
\textbf{Experimental Details.}
We instantiate REMIND on Qwen3-VL-8B. GPT-4o-mini is used to derive segment-level event cues and structured strategies, GPT-5-mini generates reflective summaries for training only, and all-MiniLM-L6-v2\footnote{https://huggingface.co/sentence-transformers/all-MiniLM-L6-v2.} serves as the text encoder, $\phi(\cdot)$ for relevance scoring. We keep the top 20\% ranked segments as $\mathcal{C}_{\mathrm{ret}}$ to form $x^{(0)}$. For salience-guided grounding, we use Top-K SAE, with $\alpha=6$, $\beta=4$, and $\lambda=0.3$.
For Progressive Reflective Alignment, we set $\tau=2.0$, $\alpha_0=\alpha_1=0.5$, $\gamma_0=\gamma_1=3.0$, and $\lambda_{\mathrm{hard}}=0.2$. More details are shown in Appendix \ref{app:exp_details}.

\textbf{Baselines.} We compare REMIND on our RefMem-Bench against several representative baselines of conversational memory ability. \circled{1} \textbf{base settings}: Qwen3-VL-8B~\citep{qwen3technicalreport2025} and NaiveRAG~\citep{lewis2020retrieval}. \circled{2} \textbf{training-free memory methods}: MemGPT~\citep{packer2023memgpt}, Mem0~\citep{chhikara2025mem0}, MemoryOS~\citep{kang2025memory}, LangMem~\citep{langmem2025}, A-Mem~\citep{xu2025mem}, LightMem~\citep{lightmem2025}, and GAM~\citep{gam2025}. Following common prior-work settings, these methods use GPT-4o-mini\footnote{GPT-4o-mini refers to \texttt{gpt-4o-mini-2024-07-18} unless otherwise specified.} as the backbone model. \circled{3} \textbf{trained methods}: MemAgent~\citep{memagent2025}, MEM1~\citep{zhou2025mem1}, Qwen3-VL-30B~\citep{qwen3technicalreport2025}, and DS-Distill-Qwen-14B~\citep{deepseekr12025}. More details and open-source implementations are provided in Appendix~\ref{app:exp_baselines}.

\begin{table*}[t!]
\setlength{\tabcolsep}{2.5pt}
\centering
\caption{\text{Main results on RefMem-Bench.} We adopt Qwen3-VL-8B as the base model. \textcolor{blue}{Blue} / \textcolor{red}{red} values denote gains/drops over it. Best and second-best results are highlighted in \sethlcolor{light-green}\hl{\textbf{bold}} and \underline{\sethlcolor{light-blue}\hl{underlined}}, respectively. ``--'' indicates that the original paper setting does not support the corresponding metric.}
\label{tab:main_result}
\renewcommand{\arraystretch}{0.95}
\setlength{\dashlinedash}{1.0pt}
\setlength{\dashlinegap}{2pt}
\resizebox{0.9\linewidth}{!}{
\begin{tabular}{lcccccccc}
\toprule
\multirow{2}{*}{\diagbox[height=5ex, width=9em]{\textbf{Method}}{\textbf{Task}}}
& \multicolumn{2}{c}{\textbf{Multi-Choice}}
& \multicolumn{2}{c}{\textbf{Single-Choice}}
& \multicolumn{4}{c}{\textbf{Direct-Answer}} \\
\cmidrule(lr){2-3} \cmidrule(lr){4-5} \cmidrule(lr){6-9}
& \textbf{\small Acc} & \textbf{\small MemR} 
& \textbf{\small Acc} & \textbf{\small MemR} 
& \textbf{\small Acc} & \textbf{\small MemR} 
& \textbf{\small B1} & \textbf{\small F1} \\
\midrule
Base Model 
& 33.2 & 45.0 & 45.0 & 37.8 & 21.1 & 37.9 & 17.8 & 21.3 \\

\textit{- w/ NaiveRAG}(\citeyear{lewis2020retrieval})
& 33.6\gain{+0.4}
& 45.9\gain{+0.9}
& 47.6\gain{+2.6}
& 38.6\gain{+0.8}
& 19.8\drop{-1.3}
& 38.7\gain{+0.8}
& 17.0\drop{-0.8}
& 19.8\drop{-1.5} \\
\arrayrulecolor{gray!60}\cdashline{1-9}\arrayrulecolor{black}
\addlinespace[0.4ex]

MemGPT(\citeyear{packer2023memgpt})
& 38.4\gain{+5.2}
& 41.4\drop{-3.6}
& 47.7\gain{+2.7}
& 39.3\gain{+1.5}
& 17.3\drop{-3.8}
& 34.3\drop{-3.6}
& 16.5\drop{-1.3}
& 19.3\drop{-2.0} \\ \addlinespace[5pt]

Mem0(\citeyear{chhikara2025mem0})
& 42.5\gain{+9.3}
& --
& 48.5\gain{+3.5}
& --
& 21.8\gain{+0.7}
& --
& 17.5\drop{-0.3}
& 19.9\drop{-1.4} \\ \addlinespace[5pt]

MemoryOS(\citeyear{kang2025memory})
& 41.2\gain{+8.0}
& \cellcolor{light-blue}\underline{48.9\gain{+3.9}}
& 53.1\gain{+8.1}
& 40.9\gain{+3.1}
& 19.0\drop{-2.1}
& 35.6\drop{-2.3}
& 18.6\gain{+0.8}
& 21.2\drop{-0.1} \\ \addlinespace[5pt]

LangMem(\citeyear{langmem2025})
& 41.2\gain{+8.0}
& --
& 57.8\gain{+12.8}
& --
& 21.1\gain{+0.0}
& --
& 19.9\gain{+2.1}
& 22.3\gain{+1.0} \\ \addlinespace[5pt]

A-Mem(\citeyear{xu2025mem})
& 46.3\gain{+13.1}
& 48.5\gain{+3.5}
& 57.4\gain{+12.4}
& 44.2\gain{+6.4}
& 22.9\gain{+1.8}
& 39.6\gain{+1.7}
& 16.4\drop{-1.4}
& 18.5\drop{-2.8} \\ \addlinespace[5pt]

LightMem(\citeyear{lightmem2025})
& \cellcolor{light-blue}\underline{48.3\gain{+15.1}}
& 48.5\gain{+3.5}
& 56.9\gain{+11.9}
& 44.1\gain{+6.3}
& 23.2\gain{+2.1}
& 39.9\gain{+2.0}
& \cellcolor{light-blue}\underline{20.1\gain{+2.3}}
& \cellcolor{light-blue}\underline{23.2\gain{+1.9}} \\ \addlinespace[5pt]

GAM(\citeyear{gam2025})
& 47.0\gain{+13.8}
& 46.2\gain{+1.2}
& 56.4\gain{+11.4}
& 41.4\gain{+3.6}
& 23.2\gain{+2.1}
& 36.8\drop{-1.1}
& 19.9\gain{+2.1}
& 23.0\gain{+1.7} \\
\arrayrulecolor{gray!60}\cdashline{1-9}\arrayrulecolor{black}
\addlinespace[0.4ex]

MemAgent-7B(\citeyear{memagent2025})
& 25.0\drop{-8.2}
& 40.6\drop{-4.4}
& 53.5\gain{+8.5}
& 43.9\gain{+6.1}
& 12.6\drop{-8.5}
& 29.7\drop{-8.2}
& 12.1\drop{-5.7}
& 14.2\drop{-7.1} \\ \addlinespace[5pt]

MEM1-7B(\citeyear{zhou2025mem1})
& 19.8\drop{-13.4}
& 33.9\drop{-11.1}
& 49.2\gain{+4.2}
& 36.9\drop{-0.9}
& 12.7\drop{-8.4}
& 24.6\drop{-13.3}
& 10.7\drop{-7.1}
& 12.9\drop{-8.4} \\ \addlinespace[5pt]

Qwen3-VL-30B(\citeyear{qwen3technicalreport2025})
& 36.3\gain{+3.1}
& 46.0\gain{+1.0}
& \cellcolor{light-blue}\underline{58.3\gain{+13.3}}
& 40.5\gain{+2.7}
& \cellcolor{light-blue}\underline{23.6\gain{+2.5}}
& \cellcolor{light-blue}\underline{40.4\gain{+2.5}}
& 19.5\gain{+1.7}
& 22.4\gain{+1.1} \\ \addlinespace[5pt]

DS-Distill-Qwen-14B(\citeyear{deepseekr12025})
& 24.6\drop{-8.6}
& 43.8\drop{-1.2}
& 54.9\gain{+9.9}
& \cellcolor{light-blue}\underline{45.7\gain{+7.9}}
& 19.1\drop{-2.0}
& 39.1\gain{+1.2}
& 15.4\drop{-2.4}
& 18.8\drop{-2.5} \\
\midrule

REMIND(Ours)
& \cellcolor{light-green}\textbf{59.4\gain{+26.2}}
& \cellcolor{light-green}\textbf{58.1\gain{+13.1}}
& \cellcolor{light-green}\textbf{66.2\gain{+21.2}}
& \cellcolor{light-green}\textbf{52.3\gain{+14.5}}
& \cellcolor{light-green}\textbf{32.9\gain{+11.8}}
& \cellcolor{light-green}\textbf{55.4\gain{+17.5}}
& \cellcolor{light-green}\textbf{27.6\gain{+9.8}}
& \cellcolor{light-green}\textbf{30.4\gain{+9.1}} \\
\bottomrule
\end{tabular}
}
\end{table*}

\subsection{Main Results}
As shown in Table~\ref{tab:main_result}, REMIND consistently achieves the best performance across all task formats and backbone settings, with gains that reflect a coherent and systematic improvement pattern rather than isolated metric increases. Notably, answer accuracy and memory recall improve simultaneously across settings, suggesting that the model is better at grounding its predictions in relevant evidence instead of relying on superficial correlations. This advantage becomes most evident in the direct-answer setting, where the model must both retrieve and express reasoning in an open-ended form. In this regime, REMIND shows the largest margins across both retrieval-sensitive and generation-sensitive metrics, with consistent gains jointly observed in MemR, B1, and F1, indicating that improvements in evidence recovery are effectively translated into higher-quality answer generation. Detailed per-dimension results and open-source implementation results are reported in Appendix~\ref{app:detailed_results}. These improvements are even more pronounced in the open-source implementation, suggesting that REMIND introduces a general inductive bias that enhances reasoning beyond backbone-specific capabilities. Overall, the results indicate that \textbf{REMIND improves not only prediction quality but also the underlying reasoning process by jointly strengthening evidence grounding and abstraction.}

\begin{figure}[!t]
  \centering
  \setlength{\abovecaptionskip}{-1pt}
\includegraphics[width=\linewidth]{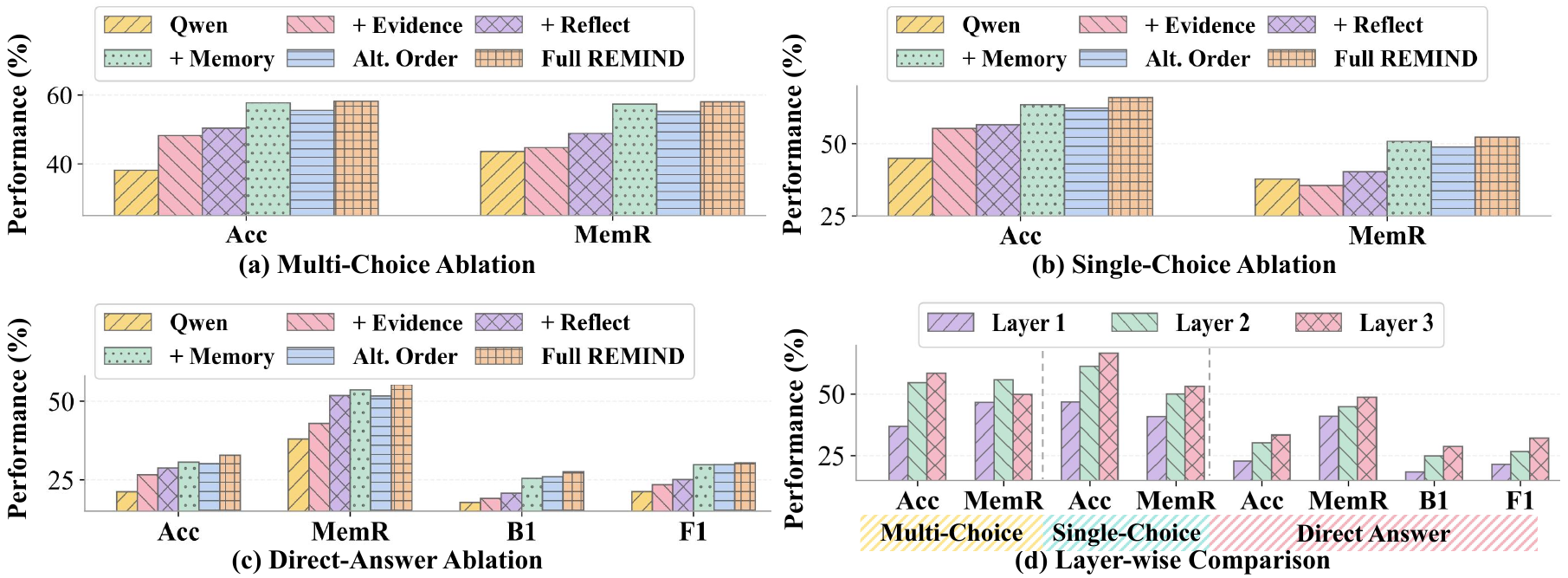}
  \caption{Results of ablation study and layer-wise analysis.
}
  \label{fig:ablation}
\vspace{-9pt}
\end{figure}

\subsection{Ablation Study and Layer Analysis}

\textbf{Ablation Study.}
To examine the contribution of each stage in REMIND, we conduct ablation experiments by progressively adding modules on top of the bare backbone model (Qwen3-VL-8B), as shown in Figure~\ref{fig:ablation} (a)(b)(c). Adding question-conditioned evidence retrieval (\textit{+ Evidence}) brings substantial gains on both Acc and MemR across all three tasks, showing that question-relevant retrieval provides a strong factual basis for reflective memory reasoning. Further introducing Reflective Strategy (\textit{+ Reflect}) improves accuracy by organizing retrieved evidence in a more question-aware way. Adding Memory Grounding (\textit{+ Memory}) yields more consistent gains, highlighting the role of the attentional level in refining coarse evidence into grounded supporting context. In contrast, \textit{Alt. Order} performs consistently worse, confirming the necessity of the proposed bottom-up hierarchy. Our REMIND achieves the best overall performance by jointly considering these components.


\textbf{Layer-wise Contribution.}
To examine whether the three levels of the Cognitive Pyramid provide progressively more useful reflective signals, we perform a layer-wise evaluation on the test set by explicitly constructing the inputs of each layer and feeding them directly into the untuned backbone model without REMIND training. As shown in Figure~\ref{fig:ablation} (d), performance improves steadily from Layer 1 to Layer 3 across all task formats, and the generation quality of direct-answer questions also increases accordingly. This stepwise trend suggests that the hierarchical construction injects increasingly useful reflective information into the backbone, rather than merely reformatting the input. Overall, the results support our core hypothesis that REMIND improves performance by progressively distilling higher-level grounding and abstraction cues into the factual inference pathway.

\subsection{Analysis of Complexity and Sensitivity}

\textbf{Complexity Analysis.}
Existing online memory systems repeatedly invoke the LLM to process the growing dialogue history, update memory, and generate the final response, so their inference cost scales with dialogue length and remains dominated by LLM-based operations. In contrast, REMIND shifts segment-level cue construction offline and reduces test-time inference to lightweight similarity matching over precomputed segment representations followed by a single answer-generation call, $C_{\text{REMIND}} = O(N)C_{\text{sim}} + O(1)C_{\text{LLM}}$.
Even with memory grounding, the additional cost is restricted to a small retrieved subset rather than the full dialogue history. This contrast shows that REMIND achieves more favorable long-context efficiency than online memory systems, which is also reflected in Figure~\ref{fig:remind-flops}. Detailed derivations are provided in Appendix~\ref{appendix:complexity}.

\begin{figure}[!t]
  \centering
\includegraphics[width=\linewidth]{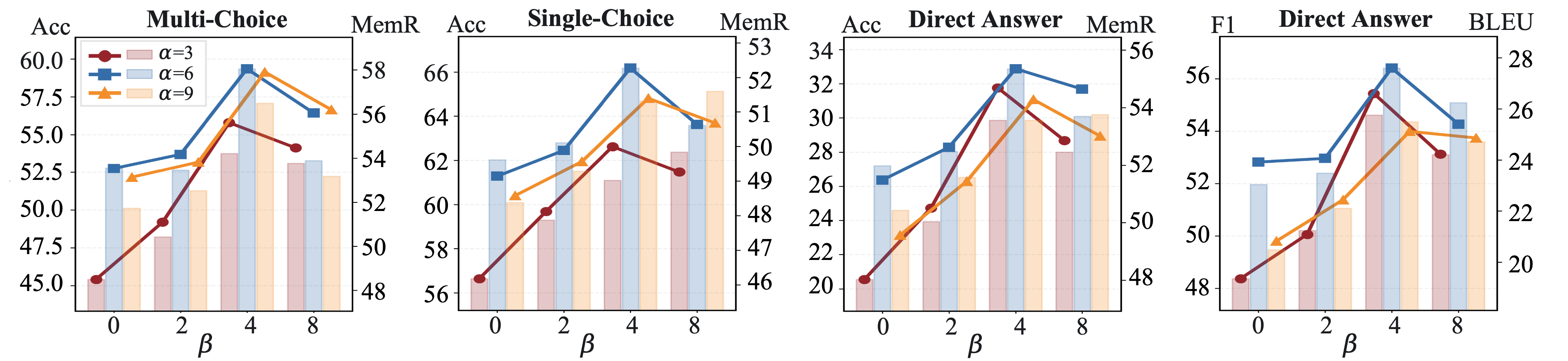}
  \caption{Sensitivity analysis for the hyperparameters $\alpha$ and $\beta$. In each subplot, lines correspond to the left-axis metric and bars correspond to the right-axis metric.
}
  \label{fig:sensitivity}
  \vspace{-6pt}
\end{figure}

\begin{figure}[!t]
  \centering
\includegraphics[width=0.91\linewidth]{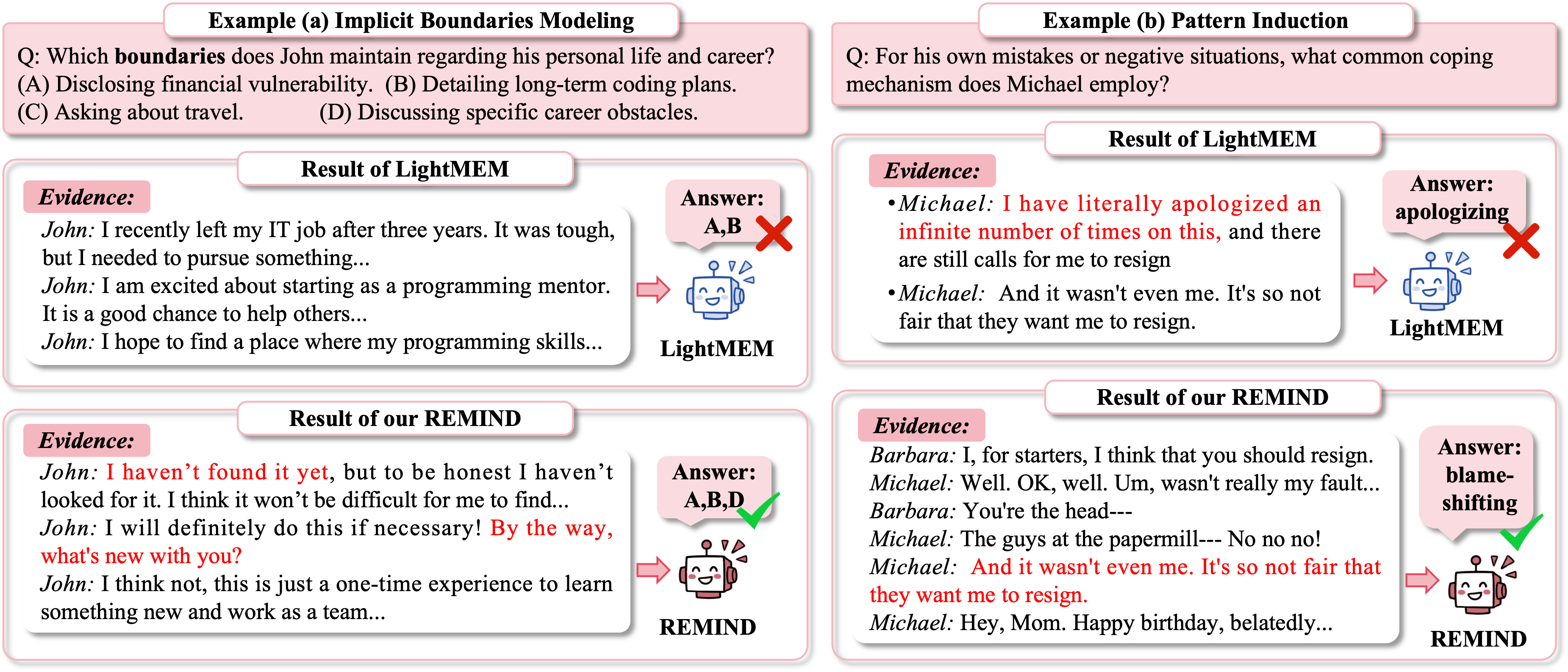}
  \caption{Two examples of the case study.}
  \label{fig:case}
\vspace{-5pt}
\end{figure}

\textbf{Sensitivity Analysis.}
We further examine the robustness of REMIND to the two key hyperparameters $\alpha$ and $\beta$ in Salience-Guided Memory Grounding.
As shown in Figure~\ref{fig:sensitivity}, three tasks exhibit a consistent trend. Performance improves as $\beta$ increases to a moderate value, but declines when it becomes large, indicating that moderate turn-level broadcasting helps recover implicit triggering context while overly strong broadcasting introduces noise and weakens evidence precision. In contrast, $\alpha$ controls the balance between explicit anchors and broader reasoning cues, with intermediate values yielding the most stable overall behavior across different $\beta$ settings. The default configuration $(\alpha, \beta) = (6, 4)$ achieves the strongest overall performance, suggesting that reflective memory reasoning benefits from a balanced combination of anchor emphasis and local structural propagation.

\subsection{Case Study}
To further illustrate the advantage of REMIND over LightMEM, Figure~\ref{fig:case} presents two representative reflective memory cases. In \textit{Implicit Boundaries Modeling}, LightMEM retrieves topically relevant but non-diagnostic evidence, whereas REMIND captures more discriminative cues such as avoidance, vulnerability suppression, and topic redirection. In \textit{Pattern Induction}, LightMEM is misled by the surface cue \emph{apologized}, while REMIND identifies the higher-level behavioral pattern of blame-shifting across turns. These examples show that REMIND is more effective at grounding reflective judgments in structurally coherent evidence and abstracting latent patterns from dispersed conversational cues.


\section{Conclusion}
Existing memory benchmarks mainly focus on explicit recall, yet real-world long-horizon dialogue often requires reflective memory: integrating temporally scattered cues into coherent higher-level interpretations. Therefore, we introduce RefMem-Bench, a benchmark for evaluating reflective memory across diverse reasoning dimensions and task formats, and propose REMIND, a hierarchical framework built on evidence perception, salience-guided grounding, and reflective abstraction. Experiments show that RefMem-Bench is challenging for current models, while REMIND consistently improves both answer accuracy and memory recall, highlighting the promise of hierarchical grounding and abstraction for LLM agents.


{
\small
\bibliographystyle{unsrtnat}
\bibliography{reference}



}

\clearpage

\appendix
\section{Appendix Overview and Organization}
This appendix provides supplementary details to support and extend the main paper. The organization of the appendix is as follows:

\begin{enumerate}
    \item \textbf{Limitations, Social Impact, and License \& Access (Section~\ref{sec:limitations}):} 
    This section discusses the limitations, broader impact, ethical considerations, licensing terms, access conditions, and reproducibility commitments of our work.

    \item \textbf{Additional Related Work (Section~\ref{sec:appendix_related_work}):} 
    This section provides further discussion of prior work on long-term memory benchmarks and long-term memory methods for LLM agents.

    \item \textbf{Dataset Construction, Annotation, and Evaluation Protocols (Section~\ref{sec:appendix_dataset}):} 
    This section details the construction of RefMem-Bench, including data sources, curation procedures, human annotation, quality control, dataset statistics, and evaluation protocols.
        
    \item \textbf{Experimental Setup, Implementation Details, and Additional Analyses (Section~\ref{sec:appendix_experiments}):} 
    This section provides supplementary experimental details, including baselines, implementation settings, detailed results, and additional analysis.
\end{enumerate}

\section{Limitations, Social Impact, and License \& Access}
\label{sec:limitations}
\subsection{Limitation}
In developing RefMem-Bench, we opted for human annotation of question-answer pairs to ensure higher data quality and richer reflective-memory scenarios. While this approach leads to a more challenging and diverse benchmark, it also introduces certain scalability constraints, as further expanding the dataset requires substantial expert effort. Additionally, during the annotation process, we intentionally avoided rigid templates to encourage greater variety and naturalness in the questions. Although this design makes the benchmark more realistic and expressive, it also makes it more challenging to precisely control the difficulty level across all instances. 

For REMIND, we design the Cognitive Pyramid to support goal-directed reasoning tasks that can be naturally organized through evidence perception, salience grounding, and higher-level abstraction. While this structure is well aligned with reflective-memory reasoning and may also benefit broader tasks such as question answering, math, personalized dialogue, adapting it to highly open-ended tasks with less explicit goals may require additional task-specific design.

Despite these inherent trade-offs, we believe that our dataset construction strategy enables RefMem-Bench to better reflect the complexity of long-horizon reflective memory, and that REMIND provides a useful framework for studying how models transform retrieved evidence into abstraction-level reasoning.

\subsection{Broader Impact}
RefMem-Bench and REMIND may bring both benefits and risks to the research community and broader society. On the positive side, RefMem-Bench provides a challenging testbed for reflective memory in long-horizon multimodal dialogue, and may support progress in personalized assistants, long-term dialogue agents, and other systems that require reliable reasoning over extended interaction histories. REMIND further supports this direction by transforming retrieved evidence into higher-level reflective reasoning.

However, stronger long-term memory capabilities may also raise privacy and fairness concerns, especially when models infer user states, preferences, or behavioral patterns from personal histories. To this end, we encourage researchers and practitioners to use RefMem-Bench and REMIND with appropriate safeguards, ethical data practices, and careful consideration of downstream deployment risks. Meanwhile, we acknowledge the computational cost of evaluating modern LLMs and MLLMs. By releasing our resources publicly, we aim to improve reproducibility and reduce redundant computation. Overall, we welcome constructive feedback and ongoing dialogue to help improve RefMem-Bench and REMIND and mitigate emerging risks.

\subsection{Ethics Statement, License and Access}
\label{app:ethics}
As this research exclusively employs publicly accessible pre-trained models and compensates annotators through biweekly or monthly salaries that cover the working hours dedicated to annotation, it does not present additional ethical concerns. We will release RefMem-Bench under the Creative Commons Attribution 4.0 International License (CC BY 4.0), and require proper disclosure when the dataset is used for model evaluation. This license does not override the original licenses of the source materials, and users must comply with all relevant legal and ethical requirements regarding data subjects. Although we make efforts to ensure the quality and legality of the benchmark, we do not guarantee absolute completeness or correctness, and assume no responsibility for legal or ethical issues arising from misuse, including copyright infringement, privacy violations, or misuse of sensitive information.

\subsection{Reproducibility Statement}
Our experiments follow baseline configurations from prior work. Implementation details are provided in the Appendix. To support reproducibility, we will release our code on GitHub .

\section{Related Work}
\label{sec:appendix_related_work}
\textbf{Additional Discussion on Long-Term Memory Benchmarks.}
Despite rapid progress in long-term memory benchmarks for LLM agents, robust evaluation of reflective memory remains limited. Most existing tasks still focus on whether models can recall, update, or apply explicitly stored information (e.g., LoCoMo~\citep{maharana2024evaluating}, MADial~\citep{he2025madial}, and LongMemEval~\citep{wu2024longmemeval}), rather than whether they can synthesize distributed evidence into higher-level interpretations. Other benchmarks, such as MemBench~\citep{tan2025membench}, MemoryAgentBench~\citep{hu2025evaluating}, MemoryBench~\citep{ai2025memorybench}, and Evo-Memory~\citep{wei2025evo}, evaluate memory from the perspective of operation and adaptation, focusing on storage, updating, continual learning, and test-time memory evolution. Furthermore, recent benchmarks such as REALTALK~\citep{lee2025realtalk}, MMRC~\citep{xue2025mmrc}, PersonaMem~\citep{jiang2025know}, KnowMe-Bench~\citep{wu2026knowme}, EMemBench~\citep{li2026emembench}, Mem-Gallery~\citep{bei2026mem}, and LifeBench~\citep{cheng2026lifebench} shift evaluation toward realistic users, personalization, and multimodality, but still provide limited support for evaluating higher-level reflective inference over distributed evidence. In contrast, RefMem-Bench is a human-curated benchmark, carefully annotated by three memory-focused experts. It provides comprehensive coverage of eight reflective-memory dimensions and three task formats, requiring models to integrate distributed evidence from extended interaction histories to infer latent user states, behavioral regularities, temporal evolution, and implicit constraints. This benchmark reveals the substantial gap between current LLM agents and human-level reflective memory, and sets a higher standard for evaluating long-horizon memory reasoning in multimodal dialogue scenarios.

\textbf{Additional Discussion on Long-Term Memory Methods.}
Existing long-term memory methods for LLM agents mainly improve long-horizon interaction by introducing external memory management, lightweight memory consolidation, or adaptive context construction. Representative systems such as MemGPT~\citep{packer2023memgpt}, A-Mem~\citep{xu2025mem}, Mem0~\citep{chhikara2025mem0}, and MemoryOS~\citep{kang2025memory}. Recent works such as LightMem~\citep{lightmem2025} and GAM~\citep{gam2025} further improve efficiency through lightweight memory consolidation and just-in-time context construction. These methods improve the scalability, organization, and efficiency of long-term memory, but still mainly treat memory as an external resource to be managed. Another line of work, such as MEM1~\citep{zhou2025mem1} and MemAgent~\citep{memagent2025}, uses reinforcement-learning-based policies to optimize memory updating and long-context reasoning, but remains focused on memory operation efficiency rather than abstraction-level reflective reasoning. In contrast, REMIND organizes reasoning through factual, attentional, and reflective states, and uses Progressive Reflective Alignment to distill high-level reflective reasoning into the factual inference pathway, enabling models to transform retrieved traces into abstraction-level reasoning while maintaining efficiency.





\section{Dataset}
\label{sec:appendix_dataset}
\subsection{Details of the Data Curation}
\label{app:dataset_curation}
\paragraph{Data Collection.}
We construct RefMem-Bench from three long-horizon conversational sources: REALTALK~\citep{lee2025realtalk}, DialSim~\citep{kim2024dialsim}, and LoCoMo~\citep{maharana2024evaluating}. These sources are chosen because they provide complementary coverage of long-range interaction phenomena. REALTALK contributes realistic conversational trajectories together with speaker-level visual sharing history, which makes it suitable for multimodal reflective-memory evaluation. DialSim and LoCoMo provide long-range dialogue structure with temporally distributed topics, recurring behavioral signals, and cross-session dependencies, which are useful for text-centric reflective-memory construction. During collection, we retain complete conversational threads rather than isolated local snippets, so that each benchmark instance can be grounded in an extended interaction history and, when available, aligned visual evidence.

\paragraph{Feature Clustering.}
After collection, we organize the source material through separate text-centric and image-centric clustering pipelines. For the text-centric branch, we first segment multi-turn conversations into fine-grained topical units, so that long conversational histories can be decomposed into semantically coherent local segments. We then group related topics with Agglomerative Clustering \citep{johnson1967hierarchical,ward1963hierarchical}, which provides a flexible way to aggregate semantically similar segments without requiring a fixed number of groups in advance. Because some clusters remain overly broad after hierarchical grouping, we further refine large clusters with K-Means \citep{macqueen1967some,lloyd1982least} to improve granularity and reduce topic mixing. For the image-centric branch, we collect speaker-level image histories from REALTALK, encode them with CLIP \citep{radford2021learning}, and cluster the resulting image representations with Agglomerative Clustering. This process yields visually coherent image groups that can be aligned with their surrounding conversational context for multimodal question construction.

\paragraph{Question Generation.}
We generate benchmark items from the clustered dialogue and image groups by focusing on reflective-memory reasoning rather than surface-level fact retrieval. In the text-centric branch, clustered dialogue segments are used to construct questions that require reasoning over temporally distributed evidence, recurring behavior, latent user attitudes, experience-conditioned future actions, and hidden interpersonal or personal boundaries. Concretely, this branch is used to generate questions for \textbf{Temporal Dynamics}, \textbf{Pattern Induction}, \textbf{Belief Modeling}, \textbf{Experience-Grounded Planning}, and \textbf{Implicit Boundary Modeling}. In the image-centric branch, clustered visual histories and their aligned dialogue context are used to generate questions that require models to integrate textual and visual signals, identify user-specific visual regularities, and abstract topic-level visual prototypes. This branch is used to construct questions for \textbf{Cross-Modal Latent Inference}, \textbf{Personalized Visual Baselines}, and \textbf{Topic-Scoped Visual Prototypes}. For direct-answer items, the generation model additionally produces a list of candidate answers to facilitate subsequent human annotation. Across both branches, each generated item is paired with the supporting evidence anchors so that the intended reasoning path remains traceable.

\paragraph{Human Annotation and Verification.}
To ensure annotation quality, three expert annotators conduct the main annotation process. Given each question and its supporting evidence anchors, the annotators provide answer annotations. For direct-answer items, annotators vote on the generated candidate answers and may additionally provide their own answers when they consider none of the candidates fully correct. They also verify whether each answer is grounded in the underlying dialogue and, when applicable, the aligned visual context. During this process, they correct or flag errors introduced by LLM-based generation. They also filter out low-quality questions and revise cases with ambiguous wording, insufficient specification, or weak grounding in the source material, resulting in the removal of 8.31\% of candidate questions.  This stage serves three purposes: (1) producing reliable answer annotations; (2) preserving traceability between each question and the evidence required to answer it; and (3) ensuring sufficient reasoning depth, diversity, and grounding. All the annotators are properly briefed with the annotation objective, and a discussion among them is conducted to resolve uncertain cases. Among items with inconsistent annotations, 5.79\% are further discussed and revised through consensus with respect to both their evidence and answers.


\subsection{Data Annotation And Quality Control Interfaces}
\label{app:annotation_interfaces}
\begin{figure}[t]
    \centering
    \includegraphics[width=\linewidth]{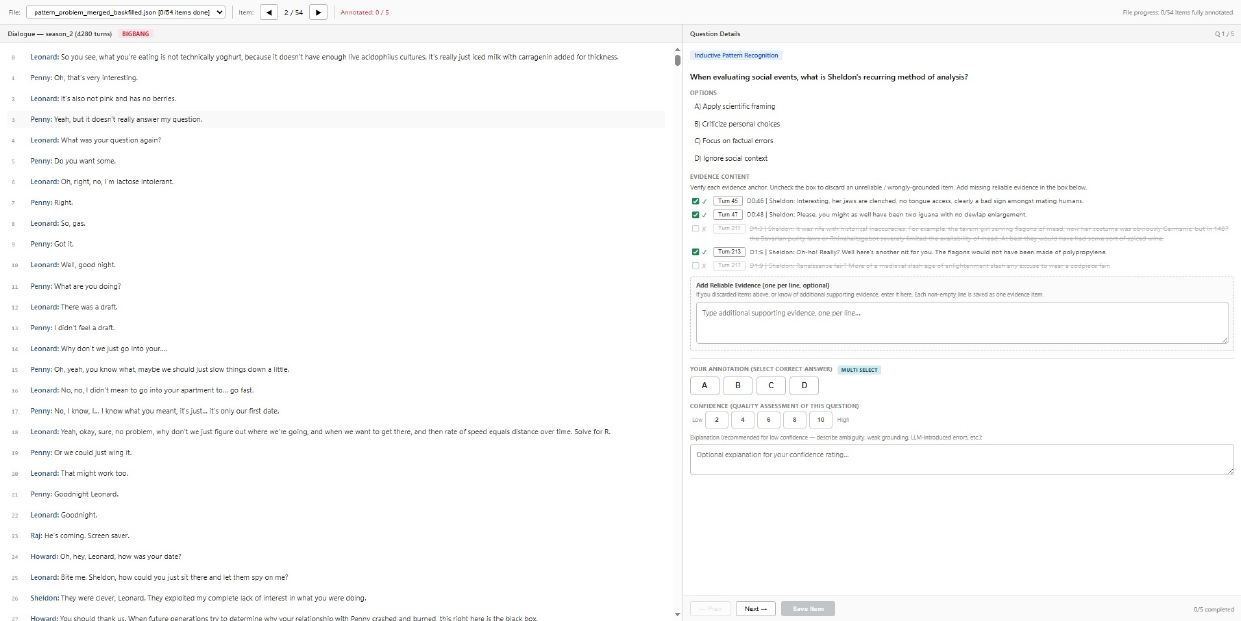}
    \caption{Screenshot of the multi-choice and single-choice annotation interface. Annotators review the dialogue context and question, select the answer option(s), verify or revise supporting evidence, and provide confidence scores for quality control.}
    \label{fig:annotation-interface-mc}
\end{figure}

\begin{figure}[ht!]
    \centering
    \includegraphics[width=\linewidth]{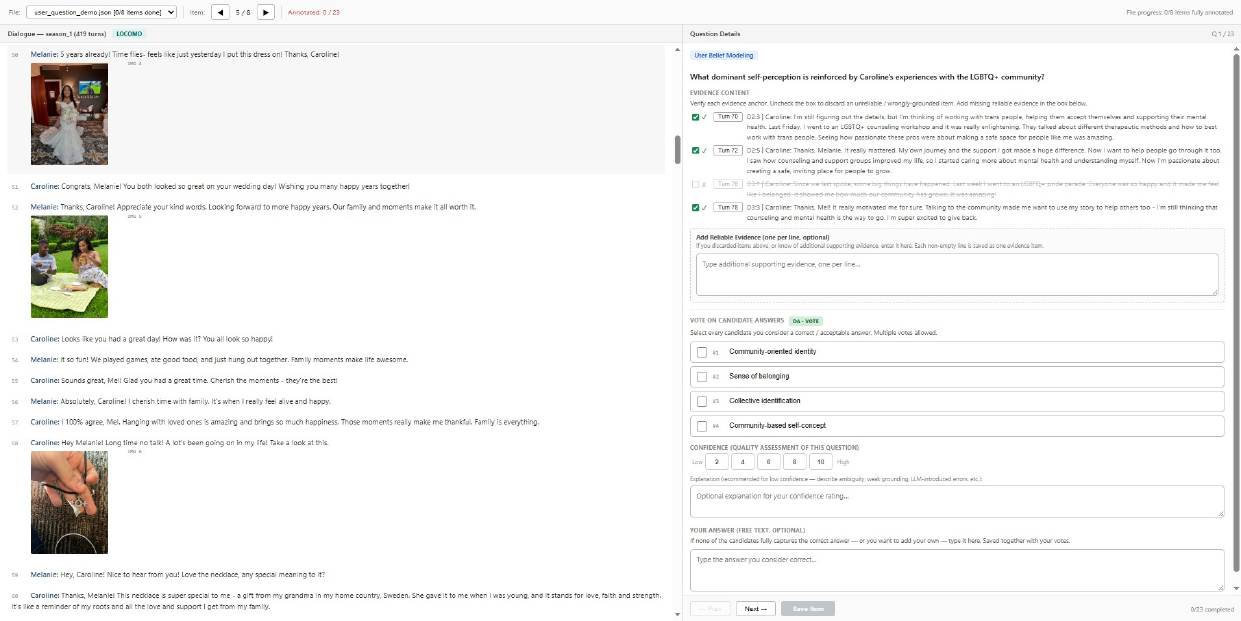}
    \caption{Screenshot of the direct-answer annotation interface. Annotators review the dialogue context and question, verify or revise supporting evidence, vote on candidate answers or provide a free-text answer, and assign confidence scores for quality control.}
    \label{fig:annotation-interface-da}
\end{figure}

The interfaces for data annotation and quality control are illustrated in Figures~\ref{fig:annotation-interface-mc} and~\ref{fig:annotation-interface-da}. For multi-choice and single-choice items, annotators review the dialogue context and question, select the correct answer option(s), verify or revise the supporting evidence, and assign confidence scores for quality control. For direct-answer items, annotators additionally vote on candidate answers or provide a free-text answer when none of the candidates is fully satisfactory.

\subsection{General Guidelines}
\label{sec:general_guidelines}

RefMem-Bench is designed to evaluate reflective-memory reasoning over long-horizon personal dialogues and aligned visual histories. The following guidelines govern the data annotation process:
\begin{itemize}
    \item \textbf{Reflective Reasoning:} All questions focus on reflective-memory reasoning rather than surface-level fact retrieval. Each item requires reasoning over long-range dialogue history, behavioral patterns, user attitudes, or experience-conditioned actions.

    \item \textbf{Dimension Coverage:} For questions in each dimension, annotators assess their quality according to the corresponding dimension definition, ensuring clear distinctions among questions from different reasoning dimensions.
    
    \item \textbf{Bias Mitigation:} The annotation interface hides the model-generated reference answer from annotators. Annotators make judgments based on the dialogue context, evidence anchors, and their own reasoning.

    \item \textbf{Difficulty, Diversity, and Clarity:} Questions must be clear, answerable by humans, and sufficiently challenging. Annotators are encouraged to remove or revise vague, repetitive, or weakly grounded items.

    \item \textbf{Multimodal Consistency:} For image-centric items, annotators verify that the visual context provides necessary evidence and that the question requires cross-modal reasoning.
\end{itemize}

\subsection{Quality Control and Validation}
\label{sec:quality_control}

A rigorous quality control procedure is implemented to ensure the reliability and validity of the benchmark:
\begin{itemize}
    \item \textbf{Independent Annotation:} All generated items are independently annotated by three in-house expert annotators with more than three years of NLP research experience.
    \item \textbf{Evidence Verification:} Annotators verify whether each evidence anchor is grounded in the underlying dialogue and, where applicable, the aligned visual context. Unreliable evidence is flagged and discarded. Missing evidence can also be supplemented to preserve the traceability of the intended reasoning path.
    \item \textbf{Quality Filtering:} Annotators remove or revise questions with ambiguous wording, insufficient specification, weak grounding, or limited reasoning depth. This step ensures that the retained questions are answerable, well grounded, and sufficiently diverse.
    \item \textbf{Inter-Annotator Agreement:} Before discussion and final resolution, we compute Fleiss' $\kappa$~\citep{fleiss1971measuring} over the three annotators' initial decisions across all question types. The resulting score, $\kappa=0.849$, indicates strong inter-annotator agreement under the interpretation of \citet{landis1977measurement}.
    \item \textbf{Disagreement Resolution:} Items with inconsistent annotations undergo structured discussion among all three annotators. Both evidence anchors and answers are jointly revised until consensus is reached.
\end{itemize}

\clearpage

\clearpage

\subsection{Reflective-Memory Dimensions}
\label{app:dimension_definitions}

\begin{figure}[!t]
  \centering
\includegraphics[width=\linewidth]{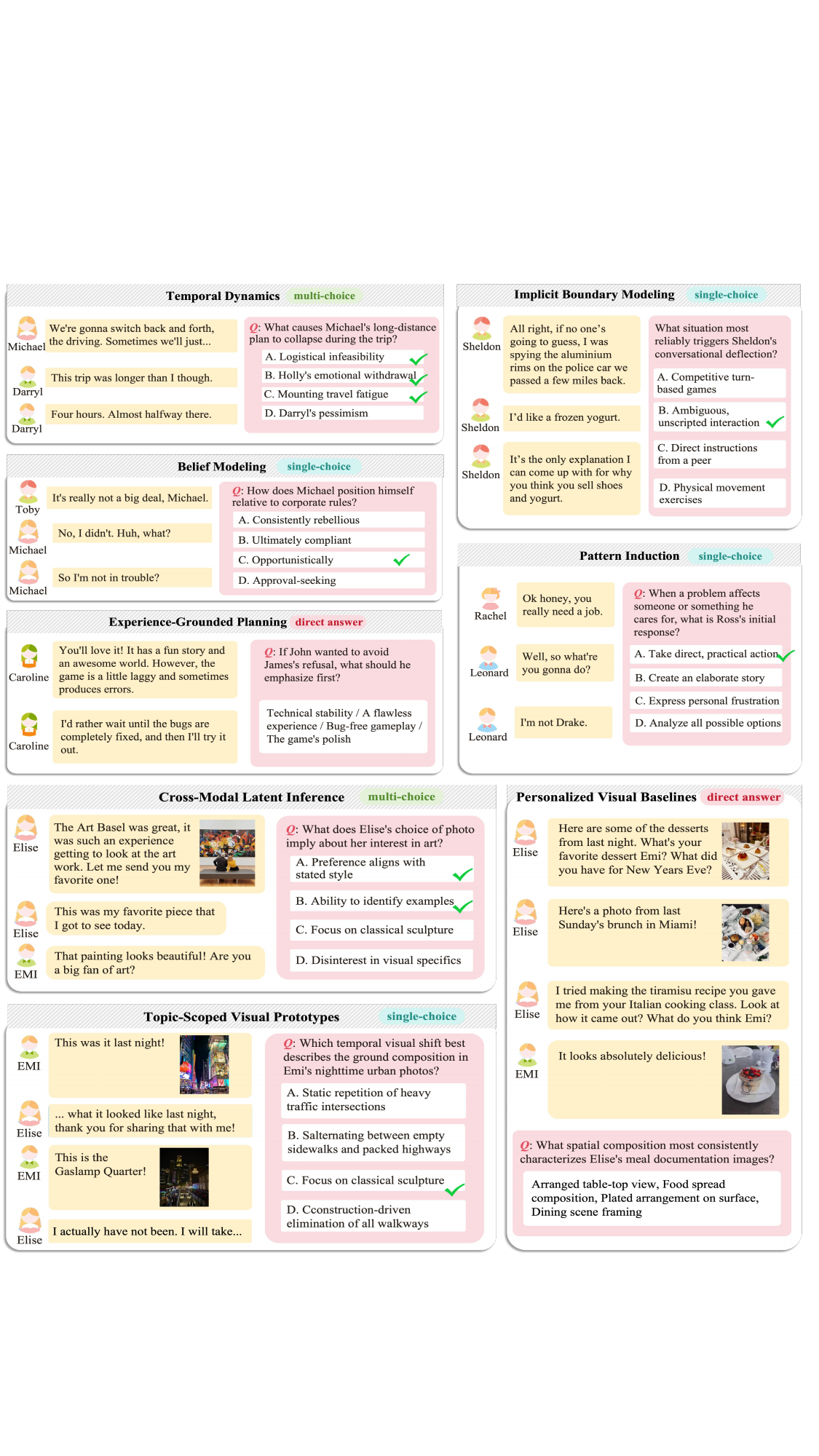}
  \caption{Visualization of our benchmark.}
  \label{fig:appendix_dataset}
\end{figure}

As shown in Figure~\ref{fig:appendix_dataset}, RefMem-Bench is organized around eight reflective-memory dimensions. These dimensions are designed to cover complementary forms of abstraction over long-horizon interactions, including temporal state tracking, latent user modeling, behavior-level regularity induction, planning from accumulated experience, and multimodal reflective reasoning.

\paragraph{Temporal Dynamics.}
This dimension evaluates whether a model can track how relevant states evolve over time. In long-horizon interactions, the correct interpretation of a user, situation, or relationship is often not determined by a single utterance or by the most recent mention alone. Instead, the model must integrate evidence distributed across turns and sessions to recover how attitudes, constraints, preferences, or circumstances shift over time. Questions in this dimension therefore test whether the model can reason over trajectories of change rather than static snapshots.

\paragraph{Pattern Induction.}
This dimension tests whether a model can abstract a stable behavioral regularity from multiple, temporally separated observations. The target is not an explicitly stated fact, but a pattern-level summary inferred from repeated instances. In many cases, the relevant evidence is scattered across distant and seemingly unrelated moments, and the answer does not appear verbatim in the source dialogue. Questions in this dimension thus assess whether the model can move from fragmented episodic evidence to a higher-level rule or recurring tendency.

\paragraph{Belief Modeling.}
This dimension evaluates whether a model can infer a coherent portrait of a user's values, preferences, decision style, or general orientation from repeated interactions. Rather than asking what the user literally said about themselves, these questions ask what kind of person the user appears to be on the basis of their accumulated behavior. The emphasis is on stable latent orientation, such as approval-seeking, risk sensitivity, or preference structure, instead of isolated profile attributes. This makes the dimension a form of user-level reflective abstraction rather than simple persona extraction.

\paragraph{Experience-Grounded Planning.}
This dimension probes whether a model can use remembered experience to infer plausible future actions, next steps, or recommendations. The answer depends on projecting a previously established behavioral tendency, preference, or practical constraint into a new or prospective situation. Unlike retrieval-style memory questions, these items require the model to convert remembered experience into an action-relevant plan or anticipatory judgment. Success in this dimension therefore depends on whether the model can ground planning in prior interaction history rather than generate generic advice.

\paragraph{Implicit Boundary Modeling.}
This dimension captures whether a model can infer hidden boundaries or sensitive triggers behind repeated avoidance, deflection, resistance, or conversational shutdown. In many dialogues, important user constraints are not explicitly declared; instead, they must be inferred from recurring negative reactions or consistent evasive behavior. Questions in this dimension therefore test whether the model can identify the latent boundary condition underlying multiple observable responses, rather than merely recalling a single avoided topic or utterance.

\paragraph{Cross-Modal Latent Inference.}
This dimension evaluates whether a model can infer an unstated capability, state, preference, or background condition by jointly reasoning over dialogue and visual evidence. Neither modality alone is sufficient to determine the answer: the target interpretation emerges only when text and image are considered together. This includes cases where the image clarifies, enriches, or partially contradicts what is said in the dialogue, thereby revealing a deeper implication not explicitly verbalized. The dimension is designed to test cross-modal reflective reasoning rather than simple image grounding.

\paragraph{Personalized Visual Baselines.}
This dimension requires the model to abstract a speaker-specific visual baseline from repeated image-sharing behavior. Instead of reasoning about a single image in isolation, the model must infer what counts as that user's typical visual style, composition, or habitual way of documenting a recurring kind of experience. The goal is to recover a personalized visual norm from longitudinal evidence. Questions in this dimension therefore test whether the model can learn stable individual-level regularities from a history of shared images.

\paragraph{Topic-Scoped Visual Prototypes.}
This dimension evaluates whether a model can distill a recurring visual prototype associated with a particular topic. Compared with Personalized Visual Baselines, which are centered on a speaker's general visual tendencies, this dimension is more topic-specific: the model must identify the characteristic visual pattern associated with a given class of content, such as meals, outings, or locations, as it appears across a user's history. The target is thus a topic-conditioned visual abstraction grounded in repeated multimodal evidence.

\subsection{Dataset Statistics of our RefMem-Bench}
\label{app:data_statistic}
Figure~\ref{fig:dataset_statistic} and Table \ref{tab:gistbench-stats} provide a more detailed view of the scale and composition of RefMem-Bench. At the context level, the benchmark contains 35 curated long-horizon conversational contexts, covering 1,341 sessions, 5,604 topic segments, and 71,062 dialogue turns in total. These contexts include 1,124 images and 958,572 words, corresponding to an average of 38.31 sessions, 2,030.34 turns, 27,387.77 words, and 32.11 images per context, with 52.99 turns per session on average. Such statistics indicate that RefMem-Bench is not only large in the number of evaluation instances, but also substantially dense in contextual evidence, making shallow recency-based strategies insufficient.

At the QA level, RefMem-Bench consists of 26,449 question-answer pairs, split into 6,170 training instances and 20,279 test instances. To reduce potential leakage, the split is conducted at the cluster-derived level rather than at the individual QA level. Specifically, all questions generated from the same semantic or visual evidence cluster are assigned entirely to either the training set or the test set. During annotation, we further manually verify the evidence grouping to avoid potential leakage from near-duplicate questions or shared local evidence across splits. After this split, the benchmark includes three task formats: 8,688 multi-choice questions, 6,812 single-choice questions, and 10,949 direct-answer questions. The dimensional distribution is broad rather than concentrated in a single reasoning type: Implicit Boundary Modeling accounts for 4,718 instances (17.8\%), Belief Modeling for 4,394 (16.6\%), Pattern Induction for 4,391 (16.6\%), Temporal Dynamics for 4,311 (16.3\%), Experience-Grounded Planning for 3,630 (13.7\%), Cross-Modal Latent Inference for 3,161 (12.0\%), Topic-Scoped Visual Prototypes for 1,050 (4.0\%), and Personalized Visual Baselines for 794 (3.0\%). In addition, 5,005 questions are derived from multimodal real-world conversations, ensuring that the benchmark tests not only text-based long-range abstraction but also image-grounded reflective reasoning. As also shown in Figure~\ref{fig:dataset_statistic}, this combination of scale, multimodality, real-world grounding, evidence anchoring, and full eight-dimension coverage distinguishes RefMem-Bench from existing dialogue resources and prior memory benchmarks.

\begin{table}[!t]
\centering
\small
\caption{Overall statistics of RefMem-Bench. We treat all sources as a unified benchmark and report aggregate conversational and QA-level statistics.}
\setlength{\tabcolsep}{6pt}
\begin{tabular}{lr}
\toprule
\multicolumn{2}{l}{\textbf{Conversation Statistics}} \\
\midrule
Organized contexts & 35 \\
Sessions & 1,341 \\
Topic segments & 5,604 \\
Dialogue turns & 71,062 \\
Images & 1,124 \\
Words & 958,572 \\
Avg. sessions / context & 38.31 \\
Avg. turns / context & 2,030.34 \\
Avg. turns / session & 52.99 \\
Avg. words / context & 27,387.77 \\
Avg. words / turn & 13.49 \\
Avg. images / context & 32.11 \\
\midrule
\multicolumn{2}{l}{\textbf{QA Benchmark Statistics}} \\
\midrule
Total questions & 26,449 \\
Multi-Choice (MC) & 8,688 \\
Single-Choice (SC) & 6,812 \\
Direct Answer (DA) & 10,949 \\
Avg. options / MC-SC item & 4.00 \\
Avg. answer cardinality & 2.71 \\
Avg. evidence anchors / item & 4.08 \\
\midrule
\multicolumn{2}{l}{\textbf{Dimension Distribution}} \\
\midrule
Temporal Dynamics & 4,311 (16.3\%) \\
Pattern Induction & 4,391 (16.6\%) \\
Belief Modeling & 4,394 (16.6\%) \\
Experience-Grounded Planning & 3,630 (13.7\%) \\
Implicit Boundary Modeling & 4,718 (17.8\%) \\
Cross-Modal Latent Inference & 3,161 (12.0\%) \\
Personalized Visual Baselines & 794 (3.0\%) \\
Topic-Scoped Visual Prototypes & 1,050 (4.0\%) \\
\bottomrule
\end{tabular}

\label{tab:gistbench-stats}
\end{table}

\subsection{Evaluation Protocols.}
\label{app:dataset_evaluation}
\begin{figure*}[!t]
    \begin{tcolorbox}
    You are a semantic equivalence judge for a direct-answer question answering task. \\
    Task: Given a question, a list of acceptable reference answers, and a model's predicted answer, determine whether the model's prediction is semantically equivalent to ANY ONE of the acceptable reference answers. \\
    Judgment Criteria: \\
    1. Semantic equivalence means the prediction and at least one reference answer refer to the same underlying concept, entity, action, or phenomenon, even if different words, synonyms, abbreviations, or paraphrases are used. \\
    2. Minor grammatical differences, such as singular versus plural, gerund versus infinitive, or articles, should not cause rejection. \\
    3. The prediction does not need to match any reference answer verbatim; conceptual alignment is sufficient. \\
    4. If the prediction refers to a different concept, a broader or narrower scope that changes the meaning, or an unrelated idea, it must be rejected. \\
    5. If the prediction is empty, nonsensical, or clearly irrelevant, reject it. \\
    Output Format: \\
    Return only a JSON object with a single boolean field ``verdict''. \\
    \{``verdict'': true\} if the prediction is semantically equivalent to any reference answer. \\
    \{``verdict'': false\} if the prediction does not match any reference answer. \\
    No extra text, no explanation, and no markdown fences. Return only the JSON object.
    \textbf{Question}: \{question\} \\
    \textbf{Acceptable Reference Answers}: \{acceptable reference answers\} \\
    \textbf{Model Prediction}: \{model's predicted answer\}
    \end{tcolorbox}
    \caption{Prompt template used by the LLM-as-a-judge for direct-answer evaluation in RefMem-Bench. The judge determines whether a model prediction is semantically equivalent to any acceptable reference answer and returns a JSON verdict only.}
    \label{fig:llm-judge-prompt}
\end{figure*}

Let $\mathcal{D}=\{(x_i,y_i)\}_{i=1}^N$ denote the evaluation set. We assess model performance with four metrics: answer accuracy, text-based memory recall, visual memory recall, and, for direct-answer questions, BLEU-1 and token-level F1. The overall answer accuracy is defined as
\begin{equation}
\mathrm{Acc}=\frac{1}{N}\sum_{i=1}^{N}s_i,
\end{equation}
where
\begin{equation}
s_i=
\begin{cases}
\mathbf{1}(P_i=G_i), & \text{for option-based tasks},\\[4pt]
J(\hat{y}_i,y_i), & \text{for direct-answer tasks},
\end{cases}
\end{equation}
with $P_i$ and $G_i$ denoting the predicted and gold option sets, respectively, $\mathbf{1}(\cdot)$ the indicator function, and $J(\hat{y}_i,y_i)\in\{0,1\}$ the LLM-as-a-judge verdict for open-ended answers. To quantify evidence recovery, we further define memory recall. For text-dominant dimensions, let $E_i^g$ and $E_i^p$ denote the gold and predicted evidence sets for sample $i$, let $\phi(\cdot)$ be the sentence encoder, and let $\cos(\cdot,\cdot)$ denote cosine similarity. The sample-level and overall text-based memory recall are defined as
\begin{equation}
\mathrm{MR}_i^{\text{text}}
=
\frac{1}{|E_i^g|}
\sum_{e\in E_i^g}
\max_{\hat e\in E_i^p}
\cos\bigl(\phi(e),\phi(\hat e)\bigr),
\qquad
\mathrm{MR}^{\text{text}}
=
\frac{1}{|\mathcal{V}|}
\sum_{i\in\mathcal{V}}
\mathrm{MR}_i^{\text{text}} .
\end{equation}
where $\mathcal{V}$ denotes the set of valid samples whose gold and predicted evidence are both non-empty. For visually grounded dimensions, let $G_i^v$ and $P_i^v$ denote the gold and predicted image-index sets for sample $i$. Similarly, the sample-level and overall visual memory recall are defined as
\begin{equation}
\mathrm{MR}_i^{\text{vis}}
=
\frac{|G_i^v\cap P_i^v|}{|G_i^v|},
\qquad
\mathrm{MR}^{\text{vis}}
=
\frac{1}{|\mathcal{V}_{\text{vis}}|}
\sum_{i\in\mathcal{V}_{\text{vis}}}
\mathrm{MR}_i^{\text{vis}} .
\end{equation}
where $\mathcal{V}_{\text{vis}}$ is the set of valid visual samples with non-empty gold image indices. For direct-answer evaluation, we additionally report BLEU-1 and token-level F1. Let $T_p$ and $T_r$ denote the tokenized prediction and reference, respectively, and let $c_p(w)$ and $c_r(w)$ denote the occurrence counts of token $w$. Defining unigram overlap as
\begin{equation}
\mathrm{Overlap}=\sum_w \min\bigl(c_p(w),c_r(w)\bigr),
\end{equation}
BLEU-1 is computed by
\begin{equation}
P_1=\frac{\mathrm{Overlap}}{|T_p|},\qquad
\mathrm{BP}=
\begin{cases}
1, & |T_p|\ge |T_r|,\\[4pt]
\exp\!\left(1-\frac{|T_r|}{|T_p|}\right), & |T_p|<|T_r|,
\end{cases}
\qquad
\mathrm{BLEU}_1=\mathrm{BP}\cdot P_1,
\end{equation}
while token-level F1 is defined as
\begin{equation}
P=\frac{\mathrm{Overlap}}{|T_p|},\qquad
R=\frac{\mathrm{Overlap}}{|T_r|},\qquad
\mathrm{F1}=\frac{2PR}{P+R}=\frac{2\cdot\mathrm{Overlap}}{|T_p|+|T_r|}.
\end{equation}

\subsection{Evaluation Metric Building}
\label{app:meta_experiments}

To accurately evaluate the diverse open-ended responses produced by different models, we use an expert-written prompt to instruct GPT-4o as the correctness judge. We present the full prompt in Figure~\ref{fig:llm-judge-prompt}. To select a reliable and cost-efficient evaluator, we conduct a meta-evaluation against expert human annotations. We consider two candidate judges, GPT-4o and Qwen3-8B-Instruct. We sample 30 questions for each dimension and report the judgment correctness by category.

As shown in Table~\ref{tab:meta_evaluation}, the prompt-engineered GPT-4o judge achieves stronger overall agreement with human annotations, with an average evaluation accuracy of 0.99 across the eight dimensions. In comparison, Qwen3-8B-Instruct obtains an average accuracy of 0.93, indicating that it is also reasonably reliable but still less aligned with expert judgments.  These results suggest that GPT-4o provides a more stable and accurate judge for our direct-answer evaluation. We will release the full evaluation prompt with the benchmark to enable consistent comparisons in future work.

\begin{table*}[h]
  \centering
  \caption{Meta-evaluation results of the prompt-engineered GPT-4o judge, showing high evaluation accuracy across all dimensions.}
  \label{tab:meta_evaluation}
  \setlength{\tabcolsep}{5pt}
  \renewcommand{\arraystretch}{1.15}
  \resizebox{0.98\textwidth}{!}{%
  \begin{tabular}{@{}lccccccccc@{}}
    \toprule
    \textbf{Models} & \textbf{TD} & \textbf{PI} & \textbf{BM} & \textbf{EGP} & \textbf{IBM} & \textbf{CMLI} & \textbf{PVB} & \textbf{TSVP} & \textbf{Avg} \\
    \midrule
    GPT-4o & 1.00 (30/30) & 1.00 (30/30) & 1.00 (30/30) & 1.00 (30/30) & 0.97 (29/30) & 0.97 (29/30) & 1.00 (30/30) & 1.00 (30/30) & 0.99 \\
    Qwen3-8B-Instruct & 0.90 (27/30) & 0.90 (27/30) & 1.00 (30/30) & 0.83 (25/30) & 1.00 (30/30) & 1.00 (30/30) & 0.90 (27/30) & 0.93 (28/30) & 0.93 \\
    \bottomrule
  \end{tabular}%
  }
\end{table*}

\section{Experiments}
\label{sec:appendix_experiments}
\subsection{Baselines} 
\label{app:exp_baselines}
We compare \textbf{REMIND} on our RefMem-Bench against several representative baselines of conversational memory ability. \circled{1} \textbf{Base settings}: Qwen3-VL-8B~\citep{qwen3technicalreport2025} and Qwen3-VL-8B with NaiveRAG~\citep{lewis2020retrieval}. \circled{2} \textbf{Training-free memory methods}: MemGPT~\citep{packer2023memgpt}, Mem0~\citep{chhikara2025mem0}, MemoryOS~\citep{kang2025memory}, LangMem~\citep{langmem2025}, A-Mem~\citep{xu2025mem}, LightMem~\citep{lightmem2025}, and GAM~\citep{gam2025}. These approaches augment a frozen backbone model with explicit memory storage, retrieval, summarization, or update mechanisms, without task-specific training on RefMem-Bench. \circled{3} \textbf{Trained methods}: MemAgent~\citep{memagent2025}, MEM1~\citep{zhou2025mem1}, Qwen3-VL-30B~\citep{qwen3technicalreport2025}, and DS-Distill-Qwen-14B~\citep{deepseekr12025}. Compared with training-free methods, these baselines either learn memory-oriented agent behaviors or benefit from stronger post-trained reasoning capabilities. This setup allows us to compare explicit memory systems against powerful trained LLMs under a unified reflective-memory evaluation protocol.For all memory baselines, we keep their memory retrieval and construction settings consistent with the original papers, including the parameter configurations and embedding models. During inference, we set the temperature to $0.0$ for all models. The prompts are provided in Appendix~\ref{appendix:prompts}.

\subsection{Implementation Details}
\label{app:exp_details}
We build REMIND on top of Qwen3-VL-8B. GPT-4o-mini is used as the auxiliary LLM to derive segment-level event cues and generate the structured strategy, while GPT-5-mini\footnote{The specific version is \texttt{gpt-5-mini-2025-08-07}.} serves as the frontier LLM and produces the reflective summaries used only during training. The text encoder $\phi(\cdot)$ underlying the relevance score $\eta_n$ is instantiated with all-MiniLM-L6-v2. We retain the top-ranked $20\%$ of segments as $\mathcal{C}_{\mathrm{ret}}$ to form the factual state $x^{(0)}$, which preserves broad evidence coverage while filtering out redundant history.

\begin{figure}[t]
    \centering
    \includegraphics[width=\linewidth]{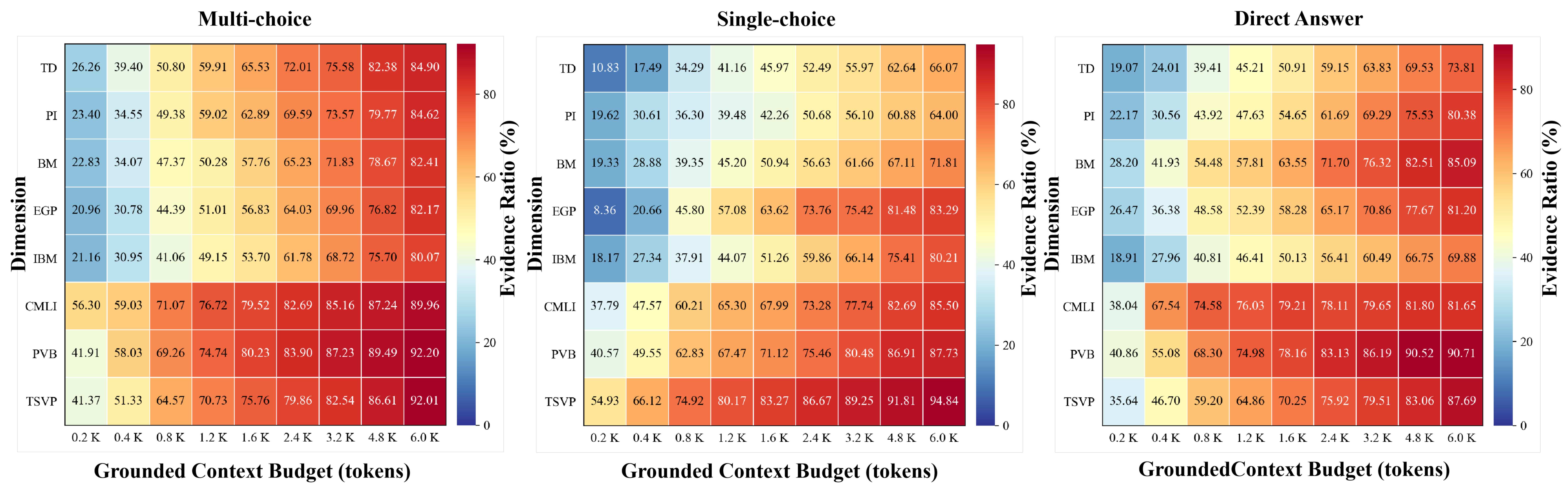}
    \caption{Evidence Ratio of salience-guided memory grounding across grounded-context budgets. The ratio measures the fraction of gold evidence covered by the strategy-salient context used to construct the attentional state.}
    \label{fig:evidence_ratio}
\end{figure}

For Top-K SAE training, we follow \citet{ma2026s}. We set the latent dimension to $d_{\mathrm{latent}}=4096$ and the sparsity parameter to $k=32$. The SAE is mounted on four representative layers of Qwen3-VL-8B. We pretrain it on DialogCC~\citep{lee2024dialogcc}, whose multi-turn multimodal dialogues match our memory-grounded setting. DialogCC shares no dialogue content with our benchmark. This alignment helps the sparse dictionary capture dialogue-specific features, including cross-turn coreference and visual anchoring. We optimize the SAE with AdamW using a learning rate of $5\times10^{-5}$. For salience-guided grounding, we build on the learned Top-K SAE sparse space. We set the anchor boosting coefficient to $\alpha=6$, the speaker-conditioned broadcast coefficient to $\beta=4$, and the modality fusion weight to $\lambda=0.3$. Each selected turn is expanded with a neighborhood window of $w=1$. Figure~\ref{fig:evidence_ratio} reports the Evidence Ratio under filtered-context budgets from $0.2$K to $6.0$K tokens. This metric measures how much gold evidence is covered by the strategy-salient context used to construct $x^{(1)}$. The consistently high ratios show that the SAE-based salience signal provides stable grounding, supporting our parameter choice and SAE training strategy. Moreover, the ratio increases rapidly up to $3.2$K and then becomes nearly saturated. We therefore set the salience threshold $\delta$ to keep a $3.2$K-token grounded context, which offers a practical trade-off between evidence coverage and context efficiency.

For progressive reflective alignment, we set the distillation temperature to $\tau=2.0$. The forward and reverse KL terms are balanced equally by $\alpha_0=\alpha_1=0.5$ for both pyramid transitions $x^{(0)}\!\rightarrow\!x^{(1)}$ and $x^{(1)}\!\rightarrow\!x^{(2)}$. The transition weights $\gamma_0=\gamma_1=3.0$ give attentional grounding and reflective abstraction comparable influence over the factual pathway. The hard--soft balance coefficient is set to $\lambda_{\mathrm{hard}}=0.2$, so that cross-level distillation carries the main learning signal while $x^{(0)}$ remains anchored to the ground-truth sequence.We optimize the model with AdamW using a learning rate of $1\times10^{-5}$. All experiments are conducted on a single node with $2\times$ NVIDIA A100 GPUs ($40$\,GB each), and the training batch size is set to $2$. All reported results are obtained with decoding temperature set to $0.0$. Detailed prompts are provided in Appendix~\ref{appendix:prompts}.

\subsection{Detailed Results on the Benchmark}
\label{app:detailed_results}
In this section, we provide dimension-wise results on RefMem-Bench. Tables~\ref{tab:main_results_openai} and~\ref{tab:direct_answer_B1_f1_openai} report the main experimental results.  Tables~\ref{tab:main_results_opensource} and~\ref{tab:direct_answer_B1_f1_opensource} report the corresponding open-source implementation results.

\begin{table*}[h]
  \centering
  \small
  \setlength{\tabcolsep}{1.5pt}
  \renewcommand{\arraystretch}{1.25}
  \setlength\dashlinedash{1.0pt}
  \setlength\dashlinegap{2pt}
  \caption{Detailed results of the main experiments on RefMem-Bench. We report answer accuracy (Acc) and memory recall (MemR) for Multi-Choice, Single-Choice, and Direct-Answer settings. ``--'' indicates that the original paper setting does not support the corresponding evaluation protocol.}
  \setlength{\tabcolsep}{2pt}
  \label{tab:main_results_openai}
  \resizebox{0.98\textwidth}{!}{%
  \begin{tabular}{l*{16}{c}}
    \toprule
    \multirow{2}{*}{\textbf{Models}}
      & \multicolumn{2}{c}{\textbf{TD}}
      & \multicolumn{2}{c}{\textbf{PI}}
      & \multicolumn{2}{c}{\textbf{BM}}
      & \multicolumn{2}{c}{\textbf{EGP}}
      & \multicolumn{2}{c}{\textbf{IBM}}
      & \multicolumn{2}{c}{\textbf{CMLI}}
      & \multicolumn{2}{c}{\textbf{PVB}}
      & \multicolumn{2}{c}{\textbf{TSVP}} \\
      \cmidrule(lr){2-3}\cmidrule(lr){4-5}\cmidrule(lr){6-7}\cmidrule(lr){8-9}\cmidrule(lr){10-11}\cmidrule(lr){12-13}\cmidrule(lr){14-15}\cmidrule(lr){16-17}
      & \small Acc & \small MemR & \small Acc & \small MemR & \small Acc & \small MemR & \small Acc & \small MemR & \small Acc & \small MemR & \small Acc & \small MemR & \small Acc & \small MemR & \small Acc & \small MemR \\
    \midrule
    \rowcolor{gptgray}
    \multicolumn{17}{c}{\rule{0pt}{2.4ex}\textbf{Multi-Choice}} \\
    \midrule
    Base Model & 36.88 & 48.78 & 47.44 & 50.86 & 42.31 & 47.43 & 24.38 & 48.62 & 25.55 & 46.54 & 24.38 & 39.56 & 27.85 & 39.24 & 36.64 & 38.66 \\
    \textit{- w/ NaiveRAG} & 37.44 & 50.45 & 47.68 & 51.32 & 40.44 & 49.93 & 23.86 & 51.89 & 27.19 & 47.54 & 25.27 & 37.84 & 28.73 & 38.34 & 38.17 & 39.70 \\
    \arrayrulecolor{gray!60}\cdashline{1-17}\arrayrulecolor{black}
    \addlinespace[0.4ex]
    MemGPT & 44.61 & 51.44 & 36.32 & 44.16 & 31.20 & 42.77 & 26.17 & 42.50 & 33.79 & 39.13 & 37.12 & 35.05 & 46.10 & 39.56 & 52.20 & 36.31 \\
    Mem0 & 38.14 & 44.19 & 33.72 & 44.11 & 25.33 & 43.72 & 28.07 & 44.76 & 31.32 & 43.06 & \underline{55.02} & -- & \textbf{63.07} & -- & \textbf{65.02} & -- \\
    MemoryOS & 42.78 & \underline{53.87} & 39.37 & 53.59 & 37.42 & 50.15 & 28.36 & \underline{53.95} & 32.14 & \underline{51.81} & 40.44 & 38.23 & 51.90 & 40.29 & 57.07 & \underline{49.40} \\
    LangMem & 45.49 & 47.17 & 39.28 & 46.42 & 37.78 & 44.34 & 34.06 & 45.55 & 36.17 & 43.29 & 40.44 & -- & 44.94 & -- & 51.35 & -- \\
    A-mem & 52.40 & 52.42 & 46.46 & \underline{54.72} & 43.26 & \textbf{50.40} & \underline{36.31} & 52.33 & \underline{37.67} & 45.60 & 48.47 & \underline{48.65} & 52.96 & 35.51 & 52.77 & 48.00 \\
    Light-mem & \underline{55.20} & 50.55 & 53.52 & 50.04 & \underline{50.78} & 49.53 & 33.74 & 51.24 & 34.40 & 48.38 & 48.20 & 46.73 & 53.33 & \underline{42.17} & 57.31 & 48.77 \\
    GAM & 53.32 & 49.93 & \underline{59.74} & 53.07 & 44.38 & 50.27 & 36.09 & 51.18 & 37.63 & 47.41 & 43.77 & 39.42 & 48.73 & 38.53 & 51.91 & 39.56 \\
    \arrayrulecolor{gray!60}\cdashline{1-17}\arrayrulecolor{black}
    \addlinespace[0.4ex]
    MemAgent-7B & 30.41 & 45.30 & 24.13 & 45.86 & 25.87 & 45.07 & 23.83 & 45.86 & 28.05 & 43.20 & 30.08 & 39.67 & 13.76 & 29.00 & 23.44 & 30.88 \\
    MEM1-7B & 22.32 & 42.04 & 26.14 & 41.50 & 26.11 & 45.72 & 12.99 & 40.11 & 14.83 & 40.14 & 24.38 & 20.28 & 13.13 & 21.59 & 18.27 & 19.51 \\
    Qwen3-VL-30B & 37.74 & 49.40 & 48.21 & 50.98 & 44.04 & 48.21 & 27.72 & 49.67 & 34.43 & 46.72 & 27.70 & 40.83 & 33.54 & 41.32 & 36.82 & 41.13 \\
    DS-Distill-Qwen-14B & 31.96 & 45.48 & 33.08 & 47.10 & 21.56 & 43.98 & 19.99 & 46.06 & 24.68 & 42.80 & 18.39 & 46.28 & 16.62 & 39.89 & 30.57 & 38.67 \\
    \arrayrulecolor{gray!60}\cdashline{1-17}\arrayrulecolor{black}
    \addlinespace[0.4ex]
    \textbf{Ours} & \textbf{60.92} & \textbf{56.06} & \textbf{65.04} & \textbf{55.52} & \textbf{64.20} & \underline{50.31} & \textbf{52.78} & \textbf{56.77} & \textbf{51.73} & \textbf{53.37} & \textbf{59.60} & \textbf{66.90} & \underline{60.09} & \textbf{63.37} & \underline{60.41} & \textbf{62.14} \\
    \midrule
    \rowcolor{qwenlav}
    \multicolumn{17}{c}{\rule{0pt}{2.4ex}\textbf{Single-Choice}} \\
    \midrule
    Base Model & 36.69 & 44.69 & 55.58 & 42.87 & 41.32 & 38.53 & 52.17 & 47.10 & 49.64 & 48.86 & 33.41 & 19.77 & 47.49 & 27.08 & 43.04 & 33.20 \\
    \textit{- w/ NaiveRAG} & 44.33 & 44.37 & 58.74 & 42.22 & 45.00 & 39.09 & 53.09 & 48.36 & 51.07 & \underline{51.24} & 33.57 & 18.65 & 49.96 & 29.97 & 45.25 & 35.19 \\
    \arrayrulecolor{gray!60}\cdashline{1-17}\arrayrulecolor{black}
    \addlinespace[0.4ex]
    MemGPT & 35.21 & 43.72 & 47.79 & 40.69 & 39.93 & 36.06 & 48.25 & 46.55 & 40.22 & 45.30 & 48.55 & 23.76 & 61.04 & 36.31 & 60.23 & 41.96 \\
    Mem0 & 43.79 & 38.83 & 55.64 & 36.22 & 48.18 & 33.80 & 59.53 & 39.68 & 63.63 & 40.60 & 34.90 & -- & 32.58 & -- & 49.62 & -- \\
    MemoryOS & 39.65 & 44.83 & 58.40 & 41.40 & 48.33 & 37.34 & 57.71 & 47.17 & 61.23 & 48.95 & 53.79 & 45.53 & 59.11 & 36.71 & 46.23 & 25.30 \\
    LangMem & 50.54 & 44.66 & 63.05 & 39.69 & 45.98 & 37.87 & 60.96 & 47.99 & 63.24 & 44.52 & 53.65 & -- & 63.18 & -- & 61.52 & -- \\
    A-mem & 49.75 & 46.89 & 60.40 & 43.98 & 43.46 & 39.93 & 61.96 & 50.10 & \underline{65.42} & 50.29 & 50.97 & 44.03 & \underline{64.72} & 40.49 & \underline{62.73} & 38.12 \\
    Light-mem & 49.43 & \underline{46.90} & 61.94 & 44.72 & 51.16 & 40.57 & 59.28 & \underline{50.87} & 56.31 & 51.26 & 51.96 & 39.56 & \textbf{64.73} & 40.42 & 60.60 & 38.52 \\
    GAM & 50.41 & 46.21 & \underline{64.91} & 44.56 & \textbf{59.49} & \textbf{46.55} & 49.65 & 47.80 & 65.33 & \underline{52.54} & 49.41 & 28.02 & 55.30 & 32.05 & 56.96 & 33.67 \\
    \arrayrulecolor{gray!60}\cdashline{1-17}\arrayrulecolor{black}
    \addlinespace[0.4ex]
    MemAgent-7B & 48.60 & 43.84 & 53.96 & 42.07 & 45.92 & 40.37 & 52.56 & 42.16 & 57.82 & 43.88 & 51.90 & 50.20 & 56.37 & \underline{45.38} & 60.60 & 42.98 \\
    MEM1-7B & 48.61 & 43.97 & 54.97 & 42.44 & 43.34 & 38.93 & 45.35 & 40.30 & 55.91 & 43.00 & 45.17 & 23.53 & 50.58 & 28.12 & 49.46 & 35.10 \\
    Qwen3-VL-30B & \underline{51.55} & 43.25 & 63.52 & 42.08 & 50.51 & 38.78 & 60.01 & 45.95 & 62.63 & 47.50 & \underline{55.38} & 30.37 & 62.25 & 36.69 & 60.38 & 39.29 \\
    DS-Distill-Qwen-14B & 47.93 & 44.80 & 57.05 & \underline{45.00} & 49.20 & \underline{43.30} & \underline{63.39} & 41.92 & 62.44 & 45.78 & 54.80 & \underline{56.65} & 52.00 & 39.00 & 52.53 & \underline{48.78} \\
    \arrayrulecolor{gray!60}\cdashline{1-17}\arrayrulecolor{black}
    \addlinespace[0.4ex]
    \textbf{Ours} & \textbf{60.97} & \textbf{50.68} & \textbf{68.26} & \textbf{46.55} & \underline{58.76} & 42.48 & \textbf{69.39} & \textbf{53.17} & \textbf{68.64} & \textbf{55.02} & \textbf{65.40} & \textbf{62.70} & 64.32 & \textbf{52.68} & \textbf{73.73} & \textbf{54.87} \\
    \midrule
    \rowcolor{glmblue}
    \multicolumn{17}{c}{\rule{0pt}{2.4ex}\textbf{Direct-Answer}} \\
    \midrule
    Base Model & 21.45 & 47.24 & 20.78 & 47.59 & 22.48 & 49.46 & 19.85 & 49.13 & 19.79 & 43.06 & 37.35 & 25.32 & 22.18 & 21.95 & 3.78 & 19.77 \\
    \textit{- w/ NaiveRAG} & 21.32 & 46.46 & 22.76 & 49.29 & 15.96 & 45.87 & 18.90 & 48.87 & 14.68 & 43.10 & 36.97 & 28.69 & 24.00 & 26.50 & 4.00 & 20.59 \\
    \arrayrulecolor{gray!60}\cdashline{1-17}\arrayrulecolor{black}
    \addlinespace[0.4ex]
    MemGPT & 15.44 & 45.19 & 14.26 & 45.66 & 18.76 & 48.27 & 15.20 & 48.15 & 17.42 & 40.06 & 38.07 & 17.22 & 15.70 & 12.57 & 3.43 & 17.06 \\
    Mem0 & 23.65 & 48.42 & 21.78 & 48.63 & 24.61 & 51.20 & 20.66 & 50.00 & 20.61 & 41.05 & 38.41 & -- & 19.08 & -- & 5.43 & -- \\
    MemoryOS & 17.72 & 46.53 & 16.98 & 46.67 & 21.54 & 49.23 & 16.75 & 49.18 & 18.53 & 41.05 & 36.75 & 19.96 & 18.14 & 14.06 & 5.31 & 18.16 \\
    LangMem & 22.08 & 48.42 & 16.71 & 48.65 & 23.51 & 50.93 & 18.21 & \underline{51.51} & 18.28 & 42.05 & 38.07 & -- & \underline{26.46} & -- & 5.36 & -- \\
    A-mem & 23.78 & \underline{50.80} & 21.14 & \underline{50.82} & 26.05 & 53.86 & 20.03 & 49.54 & \underline{22.86} & 43.89 & 39.37 & 25.17 & 23.83 & 22.98 & 6.43 & 19.89 \\
    Light-mem & \underline{25.67} & 50.24 & 21.14 & 49.47 & 25.61 & \underline{53.90} & 21.24 & 49.47 & 21.33 & 44.14 & 39.01 & 26.83 & 24.32 & 21.70 & \underline{7.31} & \underline{23.22} \\
    GAM & 21.15 & 45.30 & 22.42 & 47.54 & 24.03 & 49.70 & \underline{21.85} & 47.15 & 21.41 & 41.72 & 43.99 & 27.95 & 23.63 & 17.33 & 7.03 & 17.63 \\
    \arrayrulecolor{gray!60}\cdashline{1-17}\arrayrulecolor{black}
    \addlinespace[0.4ex]
    MemAgent-7B & 10.17 & 31.46 & 11.12 & 41.94 & 13.04 & 33.91 & 11.36 & 41.23 & 15.20 & 25.25 & 23.52 & 23.18 & 11.35 & 23.66 & 4.63 & 17.20 \\
    MEM1-7B & 11.20 & 22.47 & 15.34 & 44.50 & 10.09 & 21.18 & 10.57 & 41.49 & 13.77 & 17.89 & 23.98 & 20.88 & 12.62 & 12.99 & 3.63 & 15.44 \\
    Qwen3-VL-30B & 21.31 & 46.92 & 23.90 & 49.69 & \underline{26.45} & 51.35 & 21.17 & 50.64 & 21.49 & \underline{45.19} & \underline{44.67} & \underline{29.85} & 25.56 & 28.69 & 4.01 & 21.01 \\
    DS-Distill-Qwen-14B & 16.99 & 47.40 & \underline{29.04} & 49.20 & 13.80 & 50.45 & 18.06 & 43.45 & 13.45 & 42.33 & 33.97 & 29.10 & 24.00 & \underline{33.50} & 3.57 & 17.15 \\
    \arrayrulecolor{gray!60}\cdashline{1-17}\arrayrulecolor{black}
    \addlinespace[0.4ex]
    \textbf{Ours} & \textbf{27.21} & \textbf{53.73} & \textbf{31.49} & \textbf{52.84} & \textbf{35.68} & \textbf{57.24} & \textbf{26.82} & \textbf{53.91} & \textbf{28.59} & \textbf{48.96} & \textbf{59.05} & \textbf{54.15} & \textbf{43.50} & \textbf{63.28} & \textbf{10.59} & \textbf{58.68} \\
    \bottomrule
  \end{tabular}}
\end{table*}
\clearpage
\begin{table*}[h]
  \centering
  \caption{Detailed results on RefMem-Bench are reported using \textbf{open-source} implementations of training-free memory methods. Qwen3-VL-8B is used as the LLM backbone. We separately report answer accuracy (Acc) and memory recall (MemR) for Multi-Choice, Single-Choice, and Direct-Answer settings. ``--'' indicates that the original paper setting does not support the corresponding evaluation protocol.}
  \label{tab:main_results_opensource}
  \small
  \setlength{\tabcolsep}{1.5pt}
  \renewcommand{\arraystretch}{1.25}
  \setlength\dashlinedash{1.0pt}
  \setlength\dashlinegap{2pt}
  \setlength{\tabcolsep}{2pt}
  \resizebox{0.98\textwidth}{!}{%
  \begin{tabular}{l*{16}{c}}
    \toprule
    \multirow{2}{*}{\textbf{Models}}
      & \multicolumn{2}{c}{\textbf{TD}}
      & \multicolumn{2}{c}{\textbf{PI}}
      & \multicolumn{2}{c}{\textbf{BM}}
      & \multicolumn{2}{c}{\textbf{EGP}}
      & \multicolumn{2}{c}{\textbf{IBM}}
      & \multicolumn{2}{c}{\textbf{CMLI}}
      & \multicolumn{2}{c}{\textbf{PVB}}
      & \multicolumn{2}{c}{\textbf{TSVP}} \\
      \cmidrule(lr){2-3}\cmidrule(lr){4-5}\cmidrule(lr){6-7}\cmidrule(lr){8-9}\cmidrule(lr){10-11}\cmidrule(lr){12-13}\cmidrule(lr){14-15}\cmidrule(lr){16-17}
      & \small Acc & \small MemR & \small Acc & \small MemR & \small Acc & \small MemR & \small Acc & \small MemR & \small Acc & \small MemR & \small Acc & \small MemR & \small Acc & \small MemR & \small Acc & \small MemR \\
    \midrule
    \rowcolor{gptgray}
    \multicolumn{17}{c}{\rule{0pt}{2.4ex}\textbf{Multi-Choice}} \\
    \midrule
    Base Model & 36.88 & 48.78 & 47.44 & 50.86 & 42.31 & 47.43 & 24.38 & 48.62 & 25.55 & 46.54 & 24.38 & 39.56 & 27.85 & 39.24 & 36.64 & 38.66 \\
    \textit{- w/ NaiveRAG} & 37.44 & 50.45 & 47.68 & 51.32 & 40.44 & 49.93 & 23.86 & 51.89 & 27.19 & 47.54 & 25.27 & 37.84 & 28.73 & 38.34 & 38.17 & 39.70 \\
    \arrayrulecolor{gray!60}\cdashline{1-17}\arrayrulecolor{black}
    \addlinespace[0.4ex]
    MemGPT & 38.78 & 44.92 & 29.37 & 45.01 & 26.38 & 43.55 & 20.25 & 45.56 & 25.53 & 42.59 & 31.02 & 30.61 & 31.79 & 34.66 & 37.40 & 31.27 \\
    Mem0 & 12.60 & 37.43 & 10.75 & 39.20 & 11.84 & 39.27 & 12.67 & 39.76 & 7.86 & 34.77 & 33.24 & -- & 36.71 & -- & 33.93 & -- \\
    MemoryOS & 41.90 & 46.44 & 33.22 & 48.33 & 32.60 & 46.01 & 27.12 & 50.58 & 27.27 & 46.75 & 29.09 & 42.52 & 36.08 & 39.43 & \underline{53.50} & \underline{44.22} \\
    LangMem & 36.19 & \underline{50.53} & 34.51 & 50.69 & 35.56 & \textbf{50.53} & 26.17 & 50.52 & 22.84 & 46.63 & 26.59 & -- & 31.65 & -- & 41.98 & -- \\
    A-mem & \underline{48.59} & 49.88 & 42.25 & \underline{51.90} & 39.86 & 48.90 & \underline{33.73} & 52.16 & \underline{35.97} & \underline{47.89} & 39.33 & 27.89 & 48.73 & 29.60 & 46.56 & 26.75 \\
    Light-mem & 45.65 & 50.44 & 42.33 & 51.17 & 39.56 & 49.10 & 23.25 & \underline{52.92} & 26.01 & 46.67 & 38.23 & 30.25 & \underline{49.27} & 30.80 & 51.91 & 35.56 \\
    GAM & 45.63 & 49.52 & 42.99 & 50.35 & 36.49 & 48.52 & 33.58 & 52.32 & 32.13 & 46.44 & \underline{41.67} & 35.68 & 38.82 & 28.07 & 41.08 & 39.43 \\
    \arrayrulecolor{gray!60}\cdashline{1-17}\arrayrulecolor{black}
    \addlinespace[0.4ex]
    MemAgent-7B & 30.41 & 45.30 & 24.13 & 45.86 & 25.87 & 45.07 & 23.83 & 45.86 & 28.05 & 43.20 & 30.08 & 39.67 & 13.76 & 29.00 & 23.44 & 30.88 \\
    MEM1-7B & 22.32 & 42.04 & 26.14 & 41.50 & 26.11 & 45.72 & 12.99 & 40.11 & 14.83 & 40.14 & 24.38 & 20.28 & 13.13 & 21.59 & 18.27 & 19.51 \\
    Qwen3-VL-30B & 37.74 & 49.40 & \underline{48.21} & 50.98 & \underline{44.04} & 48.21 & 27.72 & 49.67 & 34.43 & 46.72 & 27.70 & 40.83 & 33.54 & \underline{41.32} & 36.82 & 41.13 \\
    DS-Distill-Qwen-14B & 31.96 & 45.48 & 33.08 & 47.10 & 21.56 & 43.98 & 19.99 & 46.06 & 24.68 & 42.80 & 18.39 & \underline{46.28} & 16.62 & 39.89 & 30.57 & 38.67 \\
    \arrayrulecolor{gray!60}\cdashline{1-17}\arrayrulecolor{black}
    \addlinespace[0.4ex]
    \textbf{Ours} & \textbf{54.92} & \textbf{56.06} & \textbf{64.04} & \textbf{55.52} & \textbf{63.20} & \underline{50.31} & \textbf{51.78} & \textbf{56.77} & \textbf{49.73} & \textbf{53.37} & \textbf{59.60} & \textbf{66.90} & \textbf{54.09} & \textbf{63.37} & \textbf{60.41} & \textbf{62.14} \\
    \midrule
    \rowcolor{qwenlav}
    \multicolumn{17}{c}{\rule{0pt}{2.4ex}\textbf{Single-Choice}} \\
    \midrule
    Base Model & 36.69 & 44.69 & 55.58 & 42.87 & 41.32 & 38.53 & 52.17 & 47.10 & 49.64 & 48.86 & 33.41 & 19.77 & 47.49 & 27.08 & 43.04 & 33.20 \\
    \textit{- w/ NaiveRAG} & 44.33 & 44.37 & 58.74 & 42.22 & 45.00 & 39.09 & 53.09 & 48.36 & 51.07 & \underline{51.24} & 33.57 & 18.65 & 49.96 & 29.97 & 45.25 & 35.19 \\
    \arrayrulecolor{gray!60}\cdashline{1-17}\arrayrulecolor{black}
    \addlinespace[0.4ex]
    MemGPT & 38.35 & \underline{45.94} & 53.02 & 41.58 & 37.58 & 40.94 & 48.47 & 48.76 & 46.59 & 50.85 & 47.74 & 20.73 & 52.60 & 26.07 & 51.82 & 34.70 \\
    Mem0 & 44.13 & 37.10 & 56.91 & 33.60 & 38.73 & 31.60 & 55.01 & 48.55 & 59.72 & 48.28 & 49.82 & -- & 50.97 & -- & 52.68 & -- \\
    MemoryOS & 37.01 & 43.62 & 52.30 & 40.84 & 38.34 & 39.59 & 52.62 & 47.60 & 46.79 & 44.76 & 49.33 & 30.27 & 55.92 & 38.54 & 55.67 & 43.60 \\
    LangMem & 42.62 & 44.37 & 61.30 & 41.10 & 47.00 & 36.91 & 62.85 & 47.69 & 47.00 & 45.08 & 48.18 & -- & 51.67 & -- & 53.46 & -- \\
    A-mem & 46.14 & 45.92 & 58.69 & 43.11 & 38.15 & 40.67 & 54.71 & \underline{48.88} & 52.19 & 47.80 & 52.56 & 36.71 & 54.91 & 34.95 & 59.31 & 37.11 \\
    Light-mem & 39.34 & 43.75 & 61.21 & 42.53 & 44.12 & 38.83 & 53.58 & 48.31 & 51.60 & 47.91 & 52.88 & 39.91 & 56.46 & 38.75 & 56.96 & 36.21 \\
    GAM & 43.14 & 44.98 & 59.06 & 44.29 & 41.01 & 39.78 & 53.58 & 48.46 & 52.40 & 46.96 & 45.53 & 36.45 & 50.19 & 39.50 & 55.46 & 35.43 \\
    \arrayrulecolor{gray!60}\cdashline{1-17}\arrayrulecolor{black}
    \addlinespace[0.4ex]
    MemAgent-7B & 48.60 & 43.84 & 53.96 & 42.07 & 45.92 & 40.37 & 52.56 & 42.16 & 57.82 & 43.88 & 51.90 & 50.20 & 56.37 & \underline{45.38} & \underline{60.60} & 42.98 \\
    MEM1-7B & 48.61 & 43.97 & 54.97 & 42.44 & 43.34 & 38.93 & 45.35 & 40.30 & 55.91 & 43.00 & 45.17 & 23.53 & 50.58 & 28.12 & 49.46 & 35.10 \\
    Qwen3-VL-30B & \underline{51.55} & 43.25 & \underline{63.52} & 42.08 & \underline{50.51} & 38.78 & 60.01 & 45.95 & \underline{62.63} & 47.50 & \underline{55.38} & 30.37 & \underline{62.25} & 36.69 & 60.38 & 39.29 \\
    DS-Distill-Qwen-14B & 47.93 & 44.80 & 57.05 & \underline{45.00} & 49.20 & \textbf{43.30} & \underline{63.39} & 41.92 & 62.44 & 45.78 & 54.80 & \underline{56.65} & 52.00 & 39.00 & 52.53 & \underline{48.78} \\
    \arrayrulecolor{gray!60}\cdashline{1-17}\arrayrulecolor{black}
    \addlinespace[0.4ex]
    \textbf{Ours} & \textbf{60.97} & \textbf{50.68} & \textbf{66.26} & \textbf{46.55} & \textbf{58.76} & \underline{42.48} & \textbf{69.39} & \textbf{53.17} & \textbf{68.64} & \textbf{55.02} & \textbf{65.40} & \textbf{62.70} & \textbf{64.32} & \textbf{52.68} & \textbf{73.73} & \textbf{54.87} \\
    \midrule
    \rowcolor{glmblue}
    \multicolumn{17}{c}{\rule{0pt}{2.4ex}\textbf{Direct-Answer}} \\
    \midrule
    Base Model & 21.45 & 47.24 & 20.78 & 47.59 & 22.48 & 49.46 & 19.85 & 49.13 & 19.79 & 43.06 & 37.35 & 25.32 & 22.18 & 21.95 & 3.78 & 19.77 \\
    \textit{- w/ NaiveRAG} & 21.32 & 46.46 & 22.76 & 49.29 & 15.96 & 45.87 & 18.90 & 48.87 & 14.68 & 43.10 & 36.97 & 28.69 & 24.00 & 26.50 & 4.00 & 20.59 \\
    \arrayrulecolor{gray!60}\cdashline{1-17}\arrayrulecolor{black}
    \addlinespace[0.4ex]    
    MemGPT & 12.97 & 35.04 & 10.57 & 35.81 & 13.85 & 41.29 & 10.62 & 40.46 & 10.27 & 33.06 & 36.43 & \underline{41.45} & 17.84 & 30.83 & 4.16 & 20.42 \\
    Mem0 & 7.76 & 35.42 & 11.05 & 35.62 & 13.05 & 37.26 & 12.75 & 41.11 & 11.80 & 40.05 & 38.40 & -- & \underline{25.91} & -- & 4.60 & -- \\
    MemoryOS & 15.83 & 38.20 & 15.48 & 38.66 & 19.28 & 45.81 & 13.39 & 41.62 & 15.87 & 40.22 & 32.10 & 35.69 & 18.74 & \underline{39.38} & 4.92 & \underline{24.75} \\
    LangMem & 20.65 & 48.42 & 19.33 & 48.65 & 19.96 & 50.31 & 19.95 & 48.34 & 18.21 & 43.46 & 38.58 & -- & 22.42 & -- & 3.31 & -- \\
    A-mem & \underline{23.15} & \underline{48.98} & 19.41 & \underline{50.67} & 23.56 & \underline{53.35} & 17.94 & 48.57 & 21.20 & 44.61 & 37.05 & 25.09 & 22.73 & 22.76 & 5.62 & 21.69 \\
    Light-mem & 21.89 & 47.99 & 17.14 & 48.39 & 19.67 & 50.16 & 17.36 & 46.74 & 17.06 & 41.12 & 36.89 & 22.09 & 18.39 & 20.74 & 4.29 & 18.71 \\
    GAM & 19.59 & 48.03 & 19.74 & 46.53 & 24.87 & 50.91 & 18.33 & 47.58 & 19.09 & 41.02 & 40.44 & 26.75 & 25.11 & 25.56 & \underline{6.03} & 21.24 \\
    \arrayrulecolor{gray!60}\cdashline{1-17}\arrayrulecolor{black}
    \addlinespace[0.4ex]
    MemAgent-7B & 10.17 & 31.46 & 11.12 & 41.94 & 13.04 & 33.91 & 11.36 & 41.23 & 15.20 & 25.25 & 23.52 & 23.18 & 11.35 & 23.66 & 4.63 & 17.20 \\
    MEM1-7B & 11.20 & 22.47 & 15.34 & 44.50 & 10.09 & 21.18 & 10.57 & 41.49 & 13.77 & 17.89 & 23.98 & 20.88 & 12.62 & 12.99 & 3.63 & 15.44 \\
    Qwen3-VL-30B & 21.31 & 46.92 & 23.90 & 49.69 & \underline{26.45} & 51.35 & \underline{21.17} & \underline{50.64} & \underline{21.49} & \underline{45.19} & \underline{44.67} & 29.85 & 25.56 & 28.69 & 4.01 & 21.01 \\
    DS-Distill-Qwen-14B & 16.99 & 47.40 & \underline{29.04} & 49.20 & 13.80 & 50.45 & 18.06 & 43.45 & 13.45 & 42.33 & 33.97 & 29.10 & 24.00 & 33.50 & 3.57 & 17.15 \\
    \arrayrulecolor{gray!60}\cdashline{1-17}\arrayrulecolor{black}
    \addlinespace[0.4ex]
    \textbf{Ours} & \textbf{27.21} & \textbf{53.73} & \textbf{31.49} & \textbf{52.84} & \textbf{35.68} & \textbf{57.24} & \textbf{25.82} & \textbf{53.91} & \textbf{28.59} & \textbf{48.96} & \textbf{59.05} & \textbf{54.15} & \textbf{43.50} & \textbf{63.28} & \textbf{10.59} & \textbf{58.68} \\
    \bottomrule
  \end{tabular}}
\end{table*}
\clearpage

\begin{table*}[t!]
  \centering
  \small
  \caption{Detailed Direct-Answer results for the main experiments on RefMem-Bench. We report BLEU-1 (B1) and token-level F1 (F1) for each evaluation dimension.}
  \label{tab:direct_answer_B1_f1_openai}
  \setlength{\tabcolsep}{2pt}
  \renewcommand{\arraystretch}{1.25}
  \setlength\dashlinedash{1.0pt}
  \setlength\dashlinegap{2pt}

  \resizebox{0.98\textwidth}{!}{%
  \begin{tabular}{l*{16}{c}}
    \toprule
    \multirow{2}{*}{\textbf{Models}}
      & \multicolumn{2}{c}{\textbf{TD}}
      & \multicolumn{2}{c}{\textbf{PI}}
      & \multicolumn{2}{c}{\textbf{BM}}
      & \multicolumn{2}{c}{\textbf{EGP}}
      & \multicolumn{2}{c}{\textbf{IBM}}
      & \multicolumn{2}{c}{\textbf{CMLI}}
      & \multicolumn{2}{c}{\textbf{PVB}}
      & \multicolumn{2}{c}{\textbf{TSVP}} \\
    \cmidrule(lr){2-3}
    \cmidrule(lr){4-5}
    \cmidrule(lr){6-7}
    \cmidrule(lr){8-9}
    \cmidrule(lr){10-11}
    \cmidrule(lr){12-13}
    \cmidrule(lr){14-15}
    \cmidrule(lr){16-17}
      & \small B1 & \small F1
      & \small B1 & \small F1
      & \small B1 & \small F1
      & \small B1 & \small F1
      & \small B1 & \small F1
      & \small B1 & \small F1
      & \small B1 & \small F1
      & \small B1 & \small F1 \\
    \midrule
    \rowcolor{glmblue}
    \multicolumn{17}{c}{\rule{0pt}{2.4ex}\textbf{Direct-Answer}} \\
    \midrule
    Base Model
      & 17.24 & 20.66
      & 16.19 & 19.13
      & 15.00 & 18.12
      & 22.12 & 24.04
      & 17.58 & 21.03
      & 37.38 & 41.92
      & 9.74 & 18.02
      & 3.92 & 4.23 \\
    \textit{- w/ NaiveRAG}
      & 19.25 & 16.28
      & 18.06 & 19.45
      & 15.31 & 17.71
      & 18.71 & 20.20
      & 15.81 & 19.60
      & 35.67 & 40.40
      & 10.09 & \underline{18.89}
      & 3.03 & 5.63 \\
    \arrayrulecolor{gray!60}\cdashline{1-17}\arrayrulecolor{black}
    \addlinespace[0.4ex]
    MemGPT
      & 14.80 & 16.85
      & 12.64 & 14.17
      & 14.39 & 16.42
      & 18.60 & 20.54
      & 17.70 & 20.61
      & 39.89 & 42.77
      & 9.27 & 16.85
      & 4.77 & 5.94 \\
    Mem0
      & 13.74 & 15.47
      & 15.26 & 16.96
      & 14.38 & 17.06
      & 21.33 & 23.53
      & 19.06 & 21.82
      & 40.39 & 43.11
      & 10.21 & 15.21
      & 5.76 & 6.00 \\
    MemoryOS
      & 17.47 & 19.68
      & 17.25 & 19.42
      & 20.40 & 22.27
      & 20.41 & 22.97
      & 18.36 & 21.78
      & 39.66 & 42.30
      & 10.02 & 14.63
      & 5.06 & 6.18 \\
    LangMem
      & 20.72 & 22.15
      & 19.72 & 21.25
      & \underline{20.98} & 22.96
      & 23.16 & 25.42
      & 18.54 & 20.94
      & 41.48 & 44.28
      & 9.12 & 14.99
      & 5.12 & 6.35 \\
    A-mem
      & 16.95 & 19.14
      & 14.55 & 16.60
      & 18.84 & 21.62
      & 18.91 & 21.20
      & 16.40 & 19.58
      & 30.74 & 33.18
      & 9.77 & 10.74
      & 5.22 & 6.22 \\
    Light-mem
      & \underline{21.93} & \underline{25.68}
      & 16.56 & 18.78
      & 19.81 & 22.76
      & 22.61 & 25.03
      & \underline{19.28} & \underline{22.79}
      & 42.67 & \underline{46.53}
      & \underline{11.37} & 16.95
      & 6.29 & 7.34 \\
    GAM
      & 17.70 & 20.36
      & 18.00 & 21.10
      & 20.21 & \underline{24.16}
      & \underline{24.39} & \underline{27.09}
      & 17.92 & 21.76
      & \underline{43.33} & 45.29
      & 11.01 & 16.54
      & \underline{6.95} & \underline{7.54} \\
    \arrayrulecolor{gray!60}\cdashline{1-17}\arrayrulecolor{black}
    \addlinespace[0.4ex]
    MemAgent-7B
      & 9.71 & 11.86
      & 9.34 & 10.56
      & 11.02 & 14.62
      & 14.25 & 16.53
      & 14.84 & 16.66
      & 25.93 & 29.75
      & 8.35 & 10.39
      & 3.16 & 3.28 \\
    MEM1-7B
      & 7.75 & 9.12
      & 13.52 & 15.85
      & 7.46 & 9.89
      & 9.73 & 11.58
      & 10.03 & 13.04
      & 24.22 & 27.74
      & 7.93 & 11.00
      & 4.76 & 5.21 \\
    Qwen3-VL-30B
      & 19.06 & 21.68
      & 20.01 & 22.16
      & 17.69 & 19.82
      & 23.99 & 26.53
      & 18.76 & 21.31
      & 41.83 & 44.11
      & 10.11 & 18.31
      & 4.29 & 5.21 \\
    DS-Distill-Qwen-14B
      & 15.65 & 18.32
      & \underline{20.83} & \underline{25.03}
      & 13.92 & 16.51
      & 19.49 & 21.70
      & 15.39 & 17.50
      & 25.12 & 28.60
      & 8.14 & 17.06
      & 4.52 & 5.60 \\
    \arrayrulecolor{gray!60}\cdashline{1-17}\arrayrulecolor{black}
    \addlinespace[0.4ex]
    \textbf{Ours}
      & \textbf{25.50} & \textbf{27.25}
      & \textbf{27.57} & \textbf{29.87}
      & \textbf{29.35} & \textbf{30.28}
      & \textbf{32.60} & \textbf{35.68}
      & \textbf{27.25} & \textbf{29.98}
      & \textbf{49.48} & \textbf{53.07}
      & \textbf{17.69} & \textbf{24.40}
      & \textbf{11.34} & \textbf{12.77} \\
    \bottomrule
  \end{tabular}}
\end{table*}

\begin{table*}[t!]
  \centering
  \small
  \caption{Additional Direct-Answer results on RefMem-Bench are reported using \textbf{open-source} implementations of training-free memory methods. Qwen3-VL-8B is used as the LLM backbone. We report BLEU-1 (B1) and token-level F1 (F1) for each evaluation dimension.}  
  \label{tab:direct_answer_B1_f1_opensource}
  \setlength{\tabcolsep}{2pt}
  \renewcommand{\arraystretch}{1.25}
  \setlength\dashlinedash{1.0pt}
  \setlength\dashlinegap{2pt}

  \resizebox{0.98\textwidth}{!}{%
  \begin{tabular}{l*{16}{c}}
    \toprule
    \multirow{2}{*}{\textbf{Models}}
      & \multicolumn{2}{c}{\textbf{TD}}
      & \multicolumn{2}{c}{\textbf{PI}}
      & \multicolumn{2}{c}{\textbf{BM}}
      & \multicolumn{2}{c}{\textbf{EGP}}
      & \multicolumn{2}{c}{\textbf{IBM}}
      & \multicolumn{2}{c}{\textbf{CMLI}}
      & \multicolumn{2}{c}{\textbf{PVB}}
      & \multicolumn{2}{c}{\textbf{TSVP}} \\
    \cmidrule(lr){2-3}
    \cmidrule(lr){4-5}
    \cmidrule(lr){6-7}
    \cmidrule(lr){8-9}
    \cmidrule(lr){10-11}
    \cmidrule(lr){12-13}
    \cmidrule(lr){14-15}
    \cmidrule(lr){16-17}
      & \small B1 & \small F1
      & \small B1 & \small F1
      & \small B1 & \small F1
      & \small B1 & \small F1
      & \small B1 & \small F1
      & \small B1 & \small F1
      & \small B1 & \small F1
      & \small B1 & \small F1 \\
    \midrule
    \rowcolor{glmblue}
    \multicolumn{17}{c}{\rule{0pt}{2.4ex}\textbf{Direct-Answer}} \\
    \midrule
    Base Model
      & 17.24 & 20.66
      & 16.19 & 19.13
      & 15.00 & 18.12
      & 22.12 & 24.04
      & 17.58 & 21.03
      & 37.38 & 41.92
      & 9.74 & 18.02
      & 3.92 & 4.23 \\
    \textit{- w/ NaiveRAG}
      & \underline{19.25} & 16.28
      & 18.06 & 19.45
      & 15.31 & 17.71
      & 18.71 & 20.20
      & 15.81 & 19.60
      & 35.67 & 40.40
      & 10.09 & \underline{18.89}
      & 3.03 & \underline{5.63} \\
    \arrayrulecolor{gray!60}\cdashline{1-17}\arrayrulecolor{black}
    \addlinespace[0.4ex]
    MemGPT
      & 13.03 & 14.49
      & 12.07 & 12.90
      & 12.00 & 12.98
      & 14.06 & 15.24
      & 13.79 & 15.42
      & 36.87 & 39.77
      & 9.07 & 13.82
      & 3.50 & 5.17 \\
    Mem0
      & 12.81 & 13.70
      & 10.29 & 11.20
      & 13.22 & 14.22
      & 15.09 & 16.21
      & 13.09 & 14.47
      & 38.21 & 41.19
      & 11.30 & 13.86
      & 2.39 & 3.46 \\
    MemoryOS
      & 15.02 & 16.54
      & 14.55 & 15.76
      & 14.32 & 15.64
      & 15.51 & 16.68
      & 15.84 & 17.60
      & 37.28 & 38.78
      & 10.49 & 12.37
      & 2.75 & 3.96 \\
    LangMem
      & 19.05 & 21.07
      & 16.69 & 19.24
      & \underline{19.23} & \underline{20.94}
      & 19.26 & 21.00
      & 17.91 & 17.03
      & 34.25 & 36.87
      & 9.13 & 12.51
      & 3.13 & 3.92 \\
    A-mem
      & 15.56 & 17.34
      & 11.84 & 12.83
      & 12.21 & 13.51
      & 14.29 & 15.51
      & 14.01 & 15.75
      & 24.44 & 26.44
      & 10.73 & 12.65
      & 2.77 & 3.44 \\
    Light-mem
      & 18.06 & 20.28
      & 12.22 & 13.61
      & 10.41 & 11.78
      & 15.71 & 17.21
      & 14.42 & 16.56
      & 36.25 & 39.36
      & 10.61 & 12.18
      & 2.51 & 3.03 \\
    GAM
      & 17.97 & 20.14
      & 16.28 & 18.14
      & 18.38 & 20.50
      & 20.26 & 22.08
      & 18.06 & 20.60
      & 38.45 & 42.23
      & \underline{11.90} & 15.86
      & 2.03 & 3.93 \\
    \arrayrulecolor{gray!60}\cdashline{1-17}\arrayrulecolor{black}
    \addlinespace[0.4ex]
    MemAgent-7B
      & 9.71 & 11.86
      & 9.34 & 10.56
      & 11.02 & 14.62
      & 14.25 & 16.53
      & 14.84 & 16.66
      & 25.93 & 29.75
      & 8.35 & 10.39
      & 3.16 & 3.28 \\
    MEM1-7B
      & 7.75 & 9.12
      & 13.52 & 15.85
      & 7.46 & 9.89
      & 11.73 & 13.58
      & 10.03 & 13.04
      & 24.22 & 27.74
      & 7.93 & 11.00
      & \underline{4.76} & 5.21 \\
    Qwen3-VL-30B
      & 19.06 & \underline{21.68}
      & 20.01 & 22.16
      & 17.69 & 19.82
      & \underline{23.99} & \underline{26.53}
      & \underline{18.76} & \underline{21.31}
      & \underline{41.83} & \underline{44.11}
      & 10.11 & 18.31
      & 4.29 & 5.21 \\
    DS-Distill-Qwen-14B
      & 15.65 & 18.32
      & \underline{20.83} & \underline{25.03}
      & 13.92 & 16.51
      & 19.49 & 21.70
      & 15.39 & 17.50
      & 25.12 & 28.60
      & 8.14 & 17.06
      & 4.52 & 5.60 \\
    \arrayrulecolor{gray!60}\cdashline{1-17}\arrayrulecolor{black}
    \addlinespace[0.4ex]
    \textbf{Ours}
      & \textbf{25.50} & \textbf{27.25}
      & \textbf{27.57} & \textbf{29.87}
      & \textbf{29.35} & \textbf{30.28}
      & \textbf{32.60} & \textbf{35.68}
      & \textbf{27.25} & \textbf{29.98}
      & \textbf{49.48} & \textbf{53.07}
      & \textbf{17.69} & \textbf{24.40}
      & \textbf{11.34} & \textbf{12.77} \\
    \bottomrule
  \end{tabular}}
\end{table*}
\clearpage

\subsection{Computation Complexity}
\label{appendix:complexity}
\begin{table*}[h]
\scriptsize
\centering
\caption{Complexity comparison between REMIND and other memory baselines.}
\label{tab:remind-bigO}
\resizebox{0.98\textwidth}{!}{%
\begin{tabular}{@{}lccc@{}}
\toprule
\textbf{System} & \textbf{(I) Segment \& Retrieval} & \textbf{(II) Construction} & \textbf{(III) Inference} \\
\midrule
\begin{tabular}[c]{@{}c@{}}Baselines\end{tabular}
& \begin{tabular}[c]{@{}c@{}}$O(N)\cdot\mathcal{C}_{\text{seg}}$ \\
  $+\,O(N)\cdot\mathcal{C}_{\text{LLM}}$ \\
  $+\,O(NM_1)\cdot\mathcal{C}_{\text{emb}}$\end{tabular}
& \begin{tabular}[c]{@{}c@{}}$O(NM_1)\cdot\mathcal{C}_{\text{sim}}$ \\
  $+\,O(NM_1R_1)\cdot\mathcal{C}_{\text{LLM}}$\end{tabular}
& \begin{tabular}[c]{@{}c@{}}$O(N)\cdot\mathcal{C}_{\text{LLM}}$ \\
  $+\,O(NM_1R_1)\cdot\mathcal{C}_{\text{LLM}}$ \\
  $+\,O(1)\cdot\mathcal{C}_{\text{LLM}}$\end{tabular} \\
\midrule
\begin{tabular}[c]{@{}c@{}}REMIND\end{tabular}
& \begin{tabular}[c]{@{}c@{}}$O(1)\cdot\mathcal{C}_{\text{emb}}$ \\
  $+\,O(N)\cdot\mathcal{C}_{\text{sim}}$ \\
  $+\,O(N\log k_1)$\end{tabular}
& \begin{tabular}[c]{@{}c@{}}$O(k_1 T\cdot k)\cdot\mathcal{C}_{\text{sae}}$ \\
  $+\,O(k_1 T)\cdot\mathcal{C}_{\text{LLM}}$\end{tabular}
& \begin{tabular}[c]{@{}c@{}}$O(N)\cdot\mathcal{C}_{\text{sim}}$ \\
  $+\,O(1)\cdot\mathcal{C}_{\text{LLM}}$\end{tabular} \\
\bottomrule
\end{tabular}%
}
\end{table*}

As shown in Table~\ref{tab:remind-bigO}, we analyze a long-horizon dialogue consisting of $N$ turns, each containing an average of $T$ tokens, and evaluate the cost of answering a reflective memory query $q$. In other memory baslines, the dialogue is first segmented into $N$ turns, and each turn triggers a summarization and indexing operation. Each summarization requires an LLM forward pass, and each segment is further encoded into $M_1$ memory entries for indexing:
\begin{equation}
\mathcal{C}^{\mathrm{I}}_{\mathrm{OMS}}
=
O(N)\cdot\mathcal{C}_{\text{seg}}
\;+\;
O(N)\cdot\mathcal{C}_{\text{LLM}}
\;+\;
O(NM_1)\cdot\mathcal{C}_{\text{emb}} .
\end{equation}
Each summarization then produces $M_1$ memory entries. A fraction $R_1$ of these entries trigger an action (e.g., an add, update, or delete operation), each of which requires an additional LLM call:
\begin{equation}
\mathcal{C}^{\mathrm{II}}_{\mathrm{OMS}}
=
O(NM_1)\cdot\mathcal{C}_{\text{sim}}
\;+\;
O(NM_1 R_1)\cdot\mathcal{C}_{\text{LLM}} .
\end{equation}
This results in a total inference cost of:
\begin{equation}
\mathcal{C}^{\mathrm{infer}}_{\mathrm{OMS}}
=
O(N)\cdot\mathcal{C}_{\text{LLM}}
\;+\;
O(NM_1R_1)\cdot\mathcal{C}_{\text{LLM}}
\;+\;
O(1)\cdot\mathcal{C}_{\text{LLM}}.
\end{equation}

\begin{figure}[t]
\centering
\includegraphics[width=0.67\linewidth]{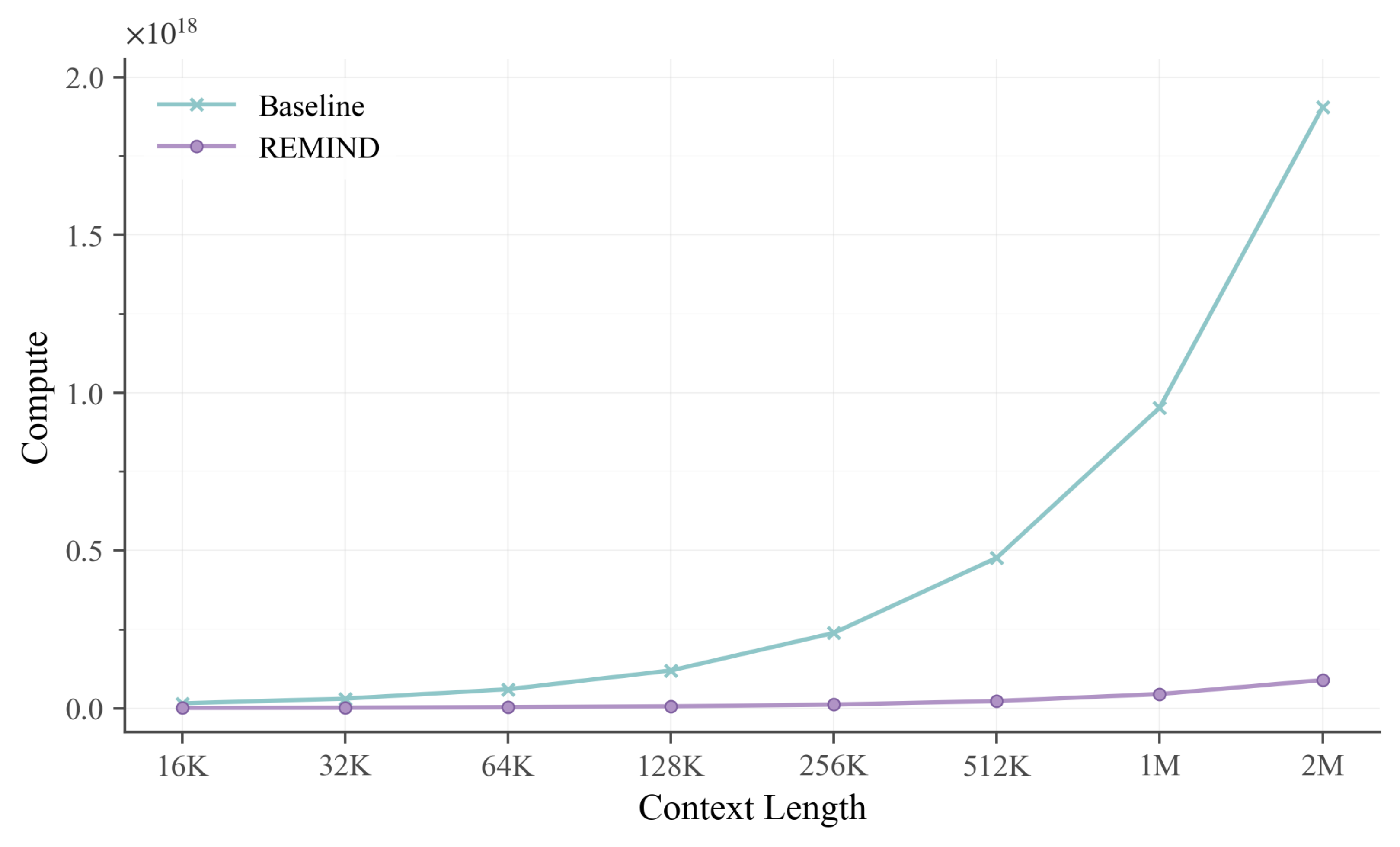}
\caption{Floating point operations across context lengths from 16K to 2M.}
\label{fig:remind-flops}

\vspace{5em}

\captionsetup{type=table}
\caption{Notation used in the complexity analysis.}
\label{tab:remind-notation}
\small
\begin{tabular}{@{}lp{0.72\linewidth}@{}}
\toprule
\textbf{Symbol} & \textbf{Definition} \\
\midrule
$N$ & Total number of dialogue turns (also the number of segments in the fact layer). \\
$T$ & Average number of tokens per turn. \\
$k_1$ & Number of segments retained after top-$k_1$ relevance selection in the fact layer. \\
$k$ & Number of active dimensions retained in Top-K SAE (sparsity parameter). \\
$M_1$ & Number of memory entries produced by each summarization in OMS. \\
$R_1$ & Fraction of memory entries that successfully retrieve at least one relevant neighbor and trigger an update in OMS. \\
$\mathcal{C}_{\text{LLM}}$ & Cost of a single-token forward pass of the backbone LLM (\emph{dominant term}); can represent runtime, FLOPs, or other computational measures. An LLM call with input length $L$ incurs cost $L \cdot \mathcal{C}_{\text{LLM}}$. \\
$\mathcal{C}_{\text{emb}}$ & Cost of a single embedding forward pass. \\
$\mathcal{C}_{\text{sim}}$ & Cost of a single vector similarity computation. \\
$\mathcal{C}_{\text{sae}}$ & Cost of a single Top-K SAE forward pass. \\
\bottomrule
\end{tabular}
\end{figure}

In \textbf{REMIND}, the segment-level cues $\{\kappa_n\}_{n=1}^{N}$ and embeddings $\{\phi(\kappa_n)\}_{n=1}^{N}$ are precomputed offline and reused across queries. Each query requires one embedding computation followed by matching against all segments:
\begin{equation}
\mathcal{C}^{\mathrm{I}}_{\mathrm{REMIND}}
=
O(1)\cdot\mathcal{C}_{\text{emb}}
\;+\;
O(N)\cdot\mathcal{C}_{\text{sim}}
\;+\;
O(N\log k_1) .
\end{equation}
On top of this, the reflective layer performs an LLM call to generate a reflective summaries. Since the input consists of $k_1$ selected segments of average length $T$, this call incurs a cost of $O(k_1 T)\cdot\mathcal{C}_{\text{LLM}}$, independent of $N$. Therefore, the total cost of Stage II in REMIND is:
\begin{equation}
\mathcal{C}^{\mathrm{II}}_{\mathrm{REMIND}}
=
O(k_1 T \cdot k)\cdot\mathcal{C}_{\text{sae}}
\;+\;
O(k_1 T)\cdot\mathcal{C}_{\text{LLM}} .
\end{equation}
The LLM cost is thus decoupled from the dialogue length $N$. Combining retrieval, attention, reflection, and final response generation, REMIND requires:
\begin{equation}
\mathcal{C}^{\mathrm{infer}}_{\mathrm{REMIND}}
=
O(N)\cdot\mathcal{C}_{\text{sim}}
\;+\;
O(1)\cdot\mathcal{C}_{\text{LLM}} .
\end{equation}
Following prior efficiency-oriented works, we report FLOPs measured by calflops \citep{calflops}. As shown in Figure~\ref{fig:remind-flops}, REMIND shifts the dominant online cost away from the expensive LLM forward pass and toward much cheaper sparse SAE computation. At the same time, through endogenous activation alignment within the cognitive pyramid, reflective reasoning is progressively absorbed into the model's internal pathway, allowing inference to rely on retrieval and the model's native capabilities rather than repeated LLM-mediated memory updates.

\clearpage
\clearpage

\subsection{Training Trajectory Analysis of REMIND}
\begin{wrapfigure}{r}{0.5\columnwidth}
    \vspace{-0.4cm}
    \centering
    \includegraphics[width=0.9\linewidth]{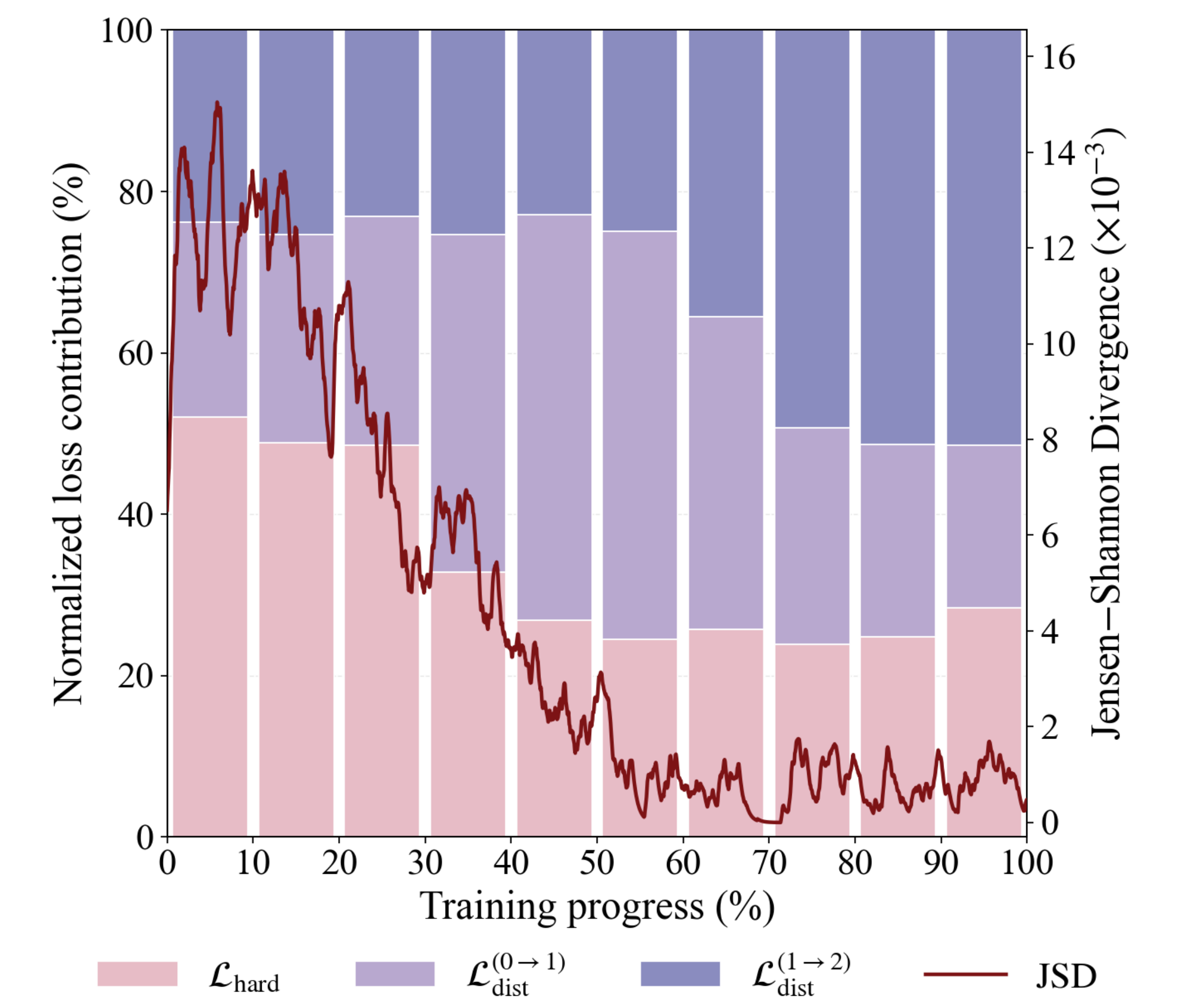}
    \caption{Training trajectory of REMIND across training progress.}
    \label{img:trajectory}
    \vspace{-0.2cm}
\end{wrapfigure}

We present the training trajectory of REMIND in Figure~\ref{img:trajectory}. 
The figure combines two complementary views of training dynamics. The stacked bars show the normalized loss distribution over equal training intervals. Each color denotes the normalized contribution of one objective term to the total loss, including the hard supervision loss $\mathcal{L}_{\mathrm{hard}}$, the lower-level distillation loss $\mathcal{L}_{\mathrm{dist}}^{(0\rightarrow 1)}$, and the higher-level distillation loss $\mathcal{L}_{\mathrm{dist}}^{(1\rightarrow 2)}$. The height of each segment therefore reflects which learning signal dominates a given training stage. The red curve reports the Jensen--Shannon divergence (JSD) between the student distribution and its direct teacher distribution. Following the Shannon-entropy formulation of Jensen--Shannon divergence introduced by \citet{Lin1991DivergenceMB}, we use its equivalent KL form to measure the discrepancy between the student and teacher output distributions:
\begin{equation} 
\mathrm{JSD}(p_s,p_t)
=
\frac{1}{2}D_{\mathrm{KL}}(p_s\|m)
+
\frac{1}{2}D_{\mathrm{KL}}(p_t\|m),
\qquad
m=\frac{1}{2}(p_s+p_t),
\end{equation}
where $p_s$ and $p_t$ denote the student and teacher output distributions, respectively. A lower JSD indicates stronger distribution-level alignment. In the early phase, covering roughly the first 30\% of training, $\mathcal{L}_{\mathrm{hard}}$ contributes the largest share of the objective. The JSD is also high and unstable. This suggests that the model still relies mainly on task-level supervision, while the student distribution has not yet aligned well with the teacher distribution.

After this initial stage, REMIND enters a middle-level alignment phase. From around 30\% to 70\% of training, $\mathcal{L}_{\mathrm{dist}}^{(0\rightarrow 1)}$ becomes the dominant signal. During the same period, the JSD decreases steadily from a high value to a near-zero level. This indicates that the attentional state begins to guide the factual pathway effectively. The model no longer only fits the hard answer label. Instead, it starts to absorb the intermediate semantic structure induced by memory grounding.

A second transition appears after around 70\% of training. The contribution of $\mathcal{L}_{\mathrm{dist}}^{(1\rightarrow 2)}$ increases and becomes dominant, while the JSD remains low. This implies that the lower-level alignment has largely stabilized. The model can then allocate more capacity to the higher-level reflective signal. At this stage, REMIND shifts from local answer fitting to reflective abstraction transfer.

Finally, in the last training phase, the loss composition becomes stable and the JSD stays close to zero. This shows that the student and teacher distributions remain well aligned after the dominant distillation signal changes. The trajectory suggests a bottom-up curriculum learned without an explicit schedule: REMIND first anchors task correctness, then absorbs attentional grounding, and finally consolidates reflective abstraction. This staged behavior supports the effectiveness of progressive reflective alignment in transferring multi-level memory reasoning into the factual inference pathway.

\clearpage
\subsection{Prompts}
\label{appendix:prompts}

\begin{table}[h]
\centering
\small
\caption{Prompt templates for text-based memory recall.}
\renewcommand{\arraystretch}{1.2}
\begin{tabularx}{\linewidth}{@{}p{2.2cm} X@{}}
\toprule
\textbf{Task type} & \textbf{Prompt template} \\
\midrule
Multi-choice
& You are an expert at long-context question answering over dialogue transcripts. Given the dialogue context and a multiple-choice question with Four options (A--D), select All correct option letters. (Note: One, two, three, or four options may be correct.) \newline
\textbf{Rules:} (1) Use Only information supported by the dialogue context. (2) Output the answer inside <answer>[A,D]</answer> -- letters must be uppercase A--D, comma-separated, sorted alphabetically. (3) Output All supporting dialogue evidence inside <explanation>...</explanation>, where each line follows the format Speaker: text\_content, copied exactly from the dialogue context. Include only relevant lines. (4) No extra text outside the tags. \newline
\textbf{Dialogue context:} \{dialogue\_context\} \newline
\textbf{Question:} \{question\} \newline
\textbf{Options:} \{options[0]\} \{options[1]\} \{options[2]\} \{options[3]\} \\
\midrule
Single-choice
& You are an expert at long-context question answering over dialogue transcripts. Given the dialogue context and a multiple-choice question with Four options (A--D), select exactly One correct option letter. (Note: Only one option is correct.) \newline
\textbf{Rules:} (1) Use Only information supported by the dialogue context. (2) Output the answer inside <answer>A</answer> -- the letter must be uppercase A--D, exactly one letter, no extra text outside the tags. (3) Output All supporting dialogue evidence inside <explanation>...</explanation>, where each line follows the format Speaker: text\_content, copied exactly from the dialogue context. Include only relevant lines. (4) No extra text outside the tags. \newline
\textbf{Dialogue context:} \{dialogue\_context\} \newline
\textbf{Question:} \{question\} \newline
\textbf{Options:} \{options[0]\} \{options[1]\} \{options[2]\} \{options[3]\} \\
\midrule
Direct answer
& You are an expert at long-context question answering over dialogue transcripts. Given the dialogue context and a direct-answer question, provide a concise answer as a short noun phrase or verb phrase of 1--4 words. \newline
\textbf{Rules:} (1) Use Only information supported by the dialogue context. (2) Output your answer inside <answer>your short phrase here</answer> -- the answer Must be a noun or verb phrase of 1--4 words, no full sentences, no ``Because ...'' explanations, use vocabulary from the dialogue when possible. (3) Output All supporting dialogue evidence inside <explanation>...</explanation>, where each line follows the format Speaker: text\_content, copied exactly from the dialogue context. Include only relevant lines. (4) No extra text outside the tags. \newline
\textbf{Dialogue context:} \{dialogue\_context\} \newline
\textbf{Question:} \{question\} \\
\bottomrule
\end{tabularx}
\label{tab:layer0_locomo}
\end{table}

\begin{table}[t]
\centering
\caption{Prompt templates for visual memory recall.}
\small
\renewcommand{\arraystretch}{1.2}
\begin{tabularx}{\linewidth}{@{}p{2.2cm} X@{}}
\toprule
\textbf{Task type} & \textbf{Prompt template} \\
\midrule
Multi-choice
& You are an expert at long-context question answering over dialogue transcripts. The dialogue context contains text turns interleaved with image captions; each caption is labeled with its 0-based global index in the format [Image \#N Caption]. Given the dialogue context and a multiple-choice question with Four options (A--D), select All correct option letters. (Note: One, two, three, or four options may be correct.) \newline
\textbf{Rules:} (1) Use Only information supported by the dialogue context. (2) Output the answer inside <answer>[A,D]</answer> -- letters must be uppercase A--D, comma-separated, sorted alphabetically. (3) Output the indices of the images you used inside <explanation>[0, 1]</explanation> as a JSON list of integers, each corresponding to the N in [Image \#N Caption]; include only relevant indices, or [] if none. (4) No extra text outside the tags. \newline
\textbf{Dialogue context:} \{dialogue\_context\} \newline
\textbf{Question:} \{question\} \newline
\textbf{Options:} \{options[0]\} \{options[1]\} \{options[2]\} \{options[3]\} \\
\midrule
Single-choice
& You are an expert at long-context question answering over dialogue transcripts. The dialogue context contains text turns interleaved with image captions; each caption is labeled with its 0-based global index in the format [Image \#N Caption]. Given the dialogue context and a multiple-choice question with Four options (A--D), select exactly One correct option letter. (Note: Only one option is correct.) \newline
\textbf{Rules:} (1) Use Only information supported by the dialogue context. (2) Output the answer inside <answer>A</answer> -- the letter must be uppercase A--D, exactly one letter, no extra text outside the tags. (3) Output the indices of the images you used inside <explanation>[0, 1]</explanation> as a JSON list of integers, each corresponding to the N in [Image \#N Caption]; include only relevant indices, or [] if none. (4) No extra text outside the tags. \newline
\textbf{Dialogue context:} \{dialogue\_context\} \newline
\textbf{Question:} \{question\} \newline
\textbf{Options:} \{options[0]\} \{options[1]\} \{options[2]\} \{options[3]\} \\
\midrule
Direct answer
& You are an expert at long-context question answering over dialogue transcripts. The dialogue context contains text turns interleaved with image captions; each caption is labeled with its 0-based global index in the format [Image \#N Caption]. Given the dialogue context and a direct-answer question, provide a concise answer as a short noun phrase or verb phrase of 1--4 words. \newline
\textbf{Rules:} (1) Use Only information supported by the dialogue context. (2) Output your answer inside <answer>your short phrase here</answer> -- the answer Must be a noun or verb phrase of 1--4 words, no full sentences, no ``Because ...'' explanations, use vocabulary from the dialogue when possible. (3) Output the indices of the images you used inside <explanation>[0, 1]</explanation> as a JSON list of integers, each corresponding to the N in [Image \#N Caption]; include only relevant indices, or [] if none. (4) No extra text outside the tags. \newline
\textbf{Dialogue context:} \{dialogue\_context\} \newline
\textbf{Question:} \{question\} \\
\bottomrule
\end{tabularx}
\label{tab:layer0_realtalk}
\end{table}

\begin{table}[t]
\centering
\caption{Level-specific prompt components across the three cognitive pyramid levels. The factual level provides the plain reconstructed dialogue with a Reflective-Memory chain-of-thought guide, the attentional level wraps the dialogue with context boundaries and soft-attention markers, and the reflective level augments the dialogue with a multi-faceted guided-reasoning analysis that must be verified against the dialogue.}
\small
\renewcommand{\arraystretch}{1.2}
\begin{tabularx}{\linewidth}{@{}p{2.4cm} X@{}}
\toprule
\textbf{Pyramid level} & \textbf{Level-specific prompt component in REMIND} \\
\midrule
Factual level
& Provides the plain reconstructed dialogue together with a Reflective-Memory chain-of-thought block that guides where to attend, but never overrides the dialogue itself. \newline
\textbf{Component:}
The following is a structured chain-of-thought analysis produced by a prior reflective memory stage. It contains anchor information (target speaker, related entities, behavioral focus, temporal scope) and step-by-step reasoning cues that help you locate and interpret the most relevant evidence in the dialogue. Use this as an analytical Guide to direct your attention when reading the dialogue context. However, you must Always verify its claims against the actual dialogue --- if the CoT contradicts what you observe in the dialogue, trust the dialogue. \newline
\{reflective\_memory\_cot\} \\
\midrule
Attentional level
& Wraps the dialogue with explicit context boundaries and injects soft-attention markers \{attention\_token\} around turns identified as semantically critical, then reinforces a marker-priority instruction. \newline
\textbf{Component:} 
\{marked\_dialogue\} \newline
\textbf{Critical reminder:} Dialogue turns enclosed by \{attention\_token\} markers contain the Most relevant evidence for answering the question below. These marked sections were identified through semantic analysis as being directly pertinent to the question. You Must prioritize information within the \{attention\_token\} markers when reasoning about the correct answer. Unmarked turns provide background context but are less likely to contain the key evidence needed for the answer. \\
\midrule
Reflective level
& Augments the dialogue with a multi-faceted guided-reasoning analysis (dimension understanding, option-grounded evidence, guiding conclusion), and explicitly demands cross-verification against the dialogue so the analysis acts as a reference rather than a verdict. \newline
\textbf{Component:} 
The following is a structured reasoning guide produced by a prior analytical stage. It contains dimension understanding, option-grounded evidence analysis, and a guiding conclusion. Use this as an analytical Reference to inform your reasoning process --- it highlights which behavioral patterns to look for and how to interpret the evidence. However, you must Always verify its claims against the actual dialogue context above. If the guided reasoning contradicts what you observe in the dialogue, trust the dialogue. \newline
\{guided\_reasoning\} \newline
\textbf{Note:} The guided reasoning above is a reference, not a definitive answer. Combine its analytical framework with the concrete evidence from the dialogue context to arrive at your final answer. Focus on verifiable behavioral patterns (repeated actions, speech patterns, avoidance moves) that you can confirm in the dialogue text. \\
\bottomrule
\end{tabularx}

\label{tab:pyramid_level_components}
\end{table}


\end{document}